\documentclass{article}

\usepackage[final]{neurips_2025}

\usepackage[utf8]{inputenc} 
\usepackage[T1]{fontenc}    
\usepackage{url}            
\usepackage{booktabs}       
\usepackage{amsfonts}       
\usepackage{nicefrac}       
\usepackage{microtype}      
\usepackage[dvipsnames]{xcolor}

\usepackage{mathtools} 
\usepackage{booktabs}
\usepackage{tikz}
\usepackage{url}
\usepackage{amssymb}
\usepackage{amsthm}
\usepackage{dsfont}
\usepackage{subcaption}
\usepackage{graphicx}
\usepackage[linesnumbered,ruled,vlined]{algorithm2e}
\usepackage{algpseudocode}
\usepackage{amsmath}  
\usepackage{mathtools}            
\usepackage{multirow}
\usepackage{array}
\usepackage{makecell}

\newtheorem{theorem}{Theorem}[section]
\newtheorem{proposition}[theorem]{Proposition}

\newtheorem{corollary}[theorem]{Corollary}
\newtheorem{definition}[theorem]{Definition}

\newtheorem*{proposition1}{Proposition~\ref{prop:mixing}}
\newtheorem*{proposition2}{Proposition~\ref{prop:convergence}}
\newtheorem*{corollary1}{Corollary~\ref{corr:loss_and_grad}}

\DeclareMathSymbol{\shortminus}{\mathbin}{AMSa}{"39}
\let\svthefootnote\thefootnote
\newcommand\blankfootnote[1]{%
  \let\thefootnote\relax\footnotetext{#1}%
  \let\thefootnote\svthefootnote%
}
\usepackage{hyperref}

\title{Learn2Mix: Training Neural Networks Using Adaptive Data Integration}

\author{%
  Shyam Venkatasubramanian \\
  Duke University\\
  \texttt{shyam.venkatasubramanian@duke.edu} \\
  \And
  Vahid Tarokh \\
  Duke University\\
  \texttt{vahid.tarokh@duke.edu}
}

\begin{document}

\maketitle
\blankfootnote{GitHub repository: \url{https://github.com/shyamven/Learn2Mix}.}
\vspace{-1.25ex}
\begin{abstract} \vspace{-0.25ex}
Accelerating model convergence in resource-constrained environments is essential for fast and efficient neural network training. This work presents \textit{learn2mix}, a new training strategy that adaptively adjusts class proportions within batches, focusing on classes with higher error rates. Unlike classical training methods that use static class proportions, learn2mix continually adapts class proportions during training, leading to faster convergence. Empirical evaluations on benchmark datasets show that neural networks trained with learn2mix converge faster than those trained with existing approaches, achieving improved results for classification, regression, and reconstruction tasks under limited training resources and with imbalanced classes. Our empirical findings are supported by theoretical analysis.
\end{abstract}

\section{Introduction}
Deep neural networks have become essential tools across various applications of machine learning, including computer vision \citep{krizhevsky2012imagenet, simonyan2014very, he2016deep}, natural language processing \citep{vaswani2017attention, devlin2018bert, radford2019language, touvron2023llamaopenefficientfoundation}, and speech recognition \citep{hinton2012deep, baevski2020wav2vec}. Despite their ability to learn and model complex, nonlinear relationships, deep neural networks often require substantial computational resources during training. In resource-constrained environments, this demand poses a significant challenge \citep{goyal2017accurate}, making the development of efficient and scalable training methodologies increasingly crucial to fully leverage the capabilities of these models.

Training deep neural networks relies on the notion of empirical risk minimization \citep{vapnik1993local}, and typically involves optimizing a loss function using gradient-based algorithms \citep{rumelhart1986learning, bottou2010large, kingma2014adam}. Techniques such as regularization \citep{srivastava2014dropout, ioffe2015batch} and data augmentation \citep{shorten2019survey}, learning rate scheduling, \citep{smith2017cyclical} and early stopping \citep{prechelt1998early}, are commonly employed to enhance generalization and prevent overfitting. However, the efficiency of the training process itself remains a critical concern, particularly in terms of convergence speed and computational resources.

Within this context, adaptive training strategies, which target enhanced generalization by modifying aspects of the training process, have emerged as promising approaches. Methods such as curriculum learning \citep{bengio2009curriculum, graves2017automated, wang2021survey} adjust the order and difficulty of training samples to facilitate more effective learning. Insights from these adaptive training strategies can be extended to the class imbalance problem \citep{wang2019dynamic}, where underrepresented classes are intrinsically harder to learn due to data scarcity \citep{buda2018systematic}, a challenge intensified in adversarial settings where safe data collection is severely limited \citep{wang2023resilient}. These methods are typically categorized into data-level methods, such as oversampling and undersampling \citep{chawla2002smote} and algorithm-level schemes, including class-balanced loss functions \citep{lin2017focal}. However, developing adaptive neural network training methodologies that \textit{accelerate} model convergence, while ensuring robustness to class imbalance, remains an open problem.

Building upon these insights, a critical aspect of training efficiency lies in the composition of batches used during stochastic gradient descent. Classical training paradigms maintain approximately fixed class proportions within each shuffled batch, mirroring the overall class distribution in the training dataset \citep{buda2018systematic, peng2019trainable}. However, this static approach fails to account for the varying levels of difficulty associated with different classes, which can hinder optimal convergence rates. For example, classes with higher error rates or those that are inherently more challenging may require greater emphasis during training to enhance model performance. While existing approaches address class imbalance by adjusting sample weights or dataset resampling, they do not dynamically change the class-wise composition of batches during training via real-time performance metrics.

\begin{figure*}[t!]
    \centering
    \captionsetup{justification=centering}
    \includegraphics[width=0.92\linewidth]{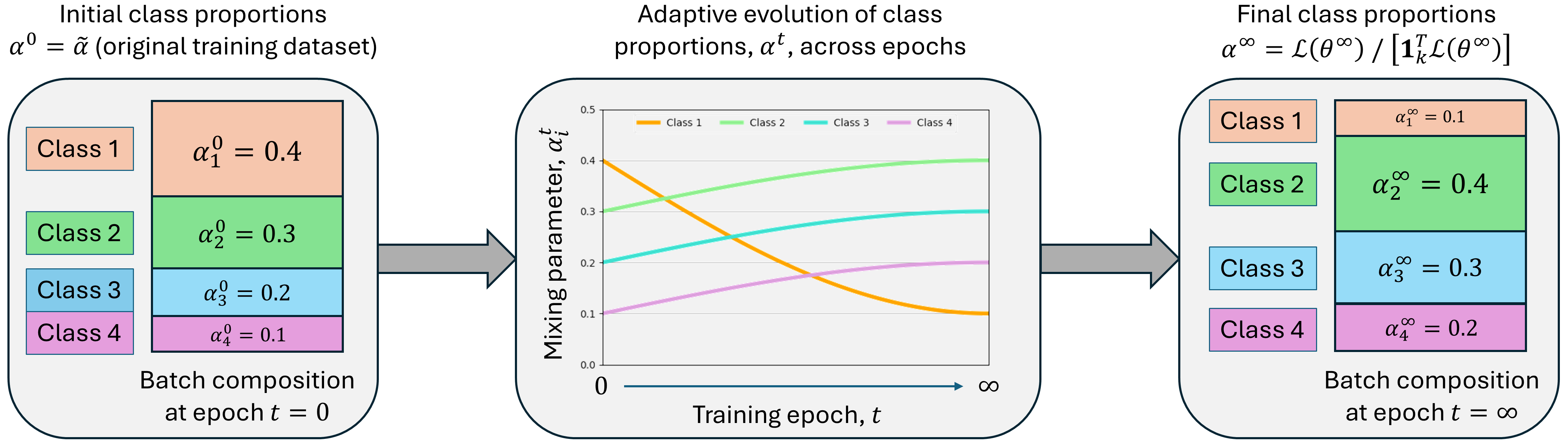}
    \caption{Illustration of the learn2mix training mechanism. The class-wise composition of batches is adaptively modified during training using instantaneous class-wise error rates.}
    \label{fig:learn2mix}
\end{figure*}

This observation motivates the central question of this paper: \textit{Can we adaptively adjust the proportion of classes within batches, across training epochs, to accelerate model convergence?} Addressing this question involves developing approaches that dynamically modify class proportions using real-time performance metrics, directing learning towards underperforming classes. Such batch construction has the potential to enhance convergence rates and training efficiency, especially in scenarios with imbalanced classes or heterogeneous class difficulties \citep{liu2008exploratory,ren2018learning}.

To address these considerations, in this work, we introduce \textit{learn2mix}, a novel training strategy that dynamically modifies class proportions in batches by emphasizing classes with higher instantaneous error rates. In contrast with classical training schemes that utilize fixed class proportions, learn2mix continually adapts these proportions during training via real-time class-wise error rates. This dynamic adjustment facilitates faster convergence and improved performance across various tasks, including classification, regression, and reconstruction. An illustration of the learn2mix training methodology is provided in Figure \ref{fig:learn2mix}, demonstrating the adaptive class-wise composition of batches.

This paper is organized as follows. In Section \ref{sec:theoretical_results}, we formalize learn2mix, and prove relevant properties. In Section \ref{sec:algorithm}, we detail the algorithmic implementation of the learn2mix training methodology. In Section \ref{sec:empirical_results}, we present empirical evaluations on benchmark datasets, demonstrating the efficacy of learn2mix in accelerating model convergence and enhancing performance. Finally, in Section \ref{sec:conclusion}, we summarize our paper. Our main contributions are outlined as follows:
\begin{enumerate}
    \item We propose \textit{learn2mix}, a novel adaptive training strategy that dynamically adjusts class proportions within batches, utilizing class-wise error rates, to accelerate model convergence.
    \item We prove that neural networks trained using \textit{learn2mix} converge faster than those trained using classical approaches when certain properties hold, such that the class proportions converge to a stable distribution proportional to the optimal class-wise error rates.
    \item We empirically validate that neural networks trained using \textit{learn2mix} consistently observe accelerated convergence, outperforming existing training methods in terms of convergence speed across classification, regression, and reconstruction tasks.
\end{enumerate}

\paragraph{Related Work.}
The landscape of neural network training methods comprises various approaches aiming to enhance model performance and training efficiency. Handling class imbalance has been extensively studied, with methods such as importance sampling \citep{katharopoulos2018not}, oversampling \citep{chawla2002smote}, undersampling \citep{tahir2012inverse}, and class-balanced loss functions \citep{lin2017focal, ren2018learning} being proposed to mitigate biases towards majority classes. In parallel, curriculum learning \citep{bengio2009curriculum} and reinforcement learning approaches \citep{florensa2017automatic} have introduced methods to facilitate more effective learning trajectories. Meta-learning, or \textit{learn2learn} methodologies \citep{arnold2020learn2learn}, including model-agnostic meta-learning (MAML) \citep{finn2017maml}, focus on optimizing the learning process itself to enable rapid adaptation to new tasks. Additionally, adaptive data sampling strategies \citep{liu2008exploratory} and boosting algorithms \citep{freund1997decision} emphasize the significance of prioritizing harder or misclassified examples to improve model robustness. Despite these advances, most existing training methods either adjust sample weights, resample datasets, or modify the sequence of training examples without specifically altering the class proportions within batches in an adaptive manner. Our proposed \textit{learn2mix} strategy distinguishes itself by adapting batch class proportions throughout the training process, targeting classes with higher error rates to accelerate convergence. This approach offers a unified framework by addressing class imbalance through adaptive training principles. 

\section{Theoretical Results} \label{sec:theoretical_results}
Consider the random variables $X \in \mathbb{R}^d$ and $Y \in \mathbb{R}^k$, where $X$ denotes the feature vector, $Y$ are the labels, and $k$ is the number of classes. We consider the \textit{original training dataset}, $J = \{(x_j, y_j)\}_{j=1}^N$, where $\smash{(x_j, y_j) \overset{\text{i.i.d.}}\sim (X, Y)}$, $\forall j \in \{1,\ldots,N \}$. The class proportions for this dataset are given by the vector of fixed-proportion mixing parameters, $\tilde{\alpha} = [\tilde{\alpha}_1, \ldots, \tilde{\alpha}_k]^T$, reflecting the distribution of classes. We define $\alpha = [\alpha_1, \ldots, \alpha_k]^T$ as a variable denoting the vector of \textit{mixing parameters}, where $\alpha_i \in [0,1]$ and $\smash{\sum_{i=1}^{\raisebox{-0.25ex}{$\scriptstyle k$}} \alpha_i = 1}$. The value of $\alpha$ determines the class proportions used during training, and can vary depending on the chosen training mechanism. In \textit{classical training}, $\alpha = \alpha^t$ is constant over time and reflects the class proportions within the original training dataset, wherein $\smash{\alpha^t = \tilde{\alpha}}$, $\forall t \in \mathbb{N}$. In \textit{learn2mix training}, $\alpha = \alpha^t$ is time-varying, and is initialized at time $t=0$ as $\alpha^0 = \tilde{\alpha}$.

Let $\mathcal{H} \subset \{h: \mathbb{R}^d \to \mathbb{R}^k\}$ be the class of hypothesis functions that model the relationship between $X$ and $Y$. For our empirical setting, we let $\mathcal{H}$ denote the set of neural networks that have predetermined architectures. We note $\mathcal{H}$ is fully defined by a vector of parameters, $\theta \in \mathbb{R}^m$, where $\mathcal{H} = h_\theta$ denotes a set of parameterized functions. The generalized form of the loss function for classical training and the loss function form under learn2mix training are given below.

\begin{definition}[Loss Function for Classical Training]
Consider $\tilde{\alpha} \in [0,1]^k$ as the vector of fixed-proportion mixing parameters, and let $\mathcal{L}(\theta^t) \in \mathbb{R}^k$ denote the vector of class-wise losses at time $t$. The loss for classical training at time $t$ is given by:
\begin{equation}
    \mathcal{L}(\theta^t, \tilde{\alpha}) = \sum_{i=1}^k \tilde{\alpha}_i \mathcal{L}_i(\theta^t) = \tilde{\alpha}^T \mathcal{L}(\theta^t).
\end{equation}
\end{definition}

\begin{definition}[Loss Function for Learn2Mix Training]
\label{def:learn2mix}
Consider $\alpha^t, \alpha^{t-1} \in [0,1]^k$ as the vector of mixing parameters at time $t$ and time $t-1$, and let $\mathcal{L}(\theta^t), \mathcal{L}(\theta^{t-1}) \in \mathbb{R}^k$ denote the respective class-wise loss vectors at time $t$ and time $t-1$. Consider $\gamma \in (0,1)$ as the mixing rate. The loss for learn2mix training at time $t$ is given by the following:
\begin{align}
    &\mathcal{L}(\theta^t, \alpha^t) = \sum_{i=1}^k \alpha^t_i \mathcal{L}_i(\theta^t) = (\alpha^t)^T \mathcal{L}(\theta^t), 
    \\ &\text{Where: } \alpha^t = \alpha^{t-1} + \gamma \bigg(\frac{\mathcal{L}(\theta^{t-1})}{\mathds{1}_k^T \mathcal{L}(\theta^{t-1})} - \alpha^{t-1} \bigg),
\end{align}
\end{definition}
We note that the denominator, $\mathds{1}_k^T \mathcal{L}(\theta^{t-1})$, is the sum of losses across all classes, and dividing by it converts $\mathcal{L}(\theta^{t-1})$ into a probability distribution. We update $\alpha^{t-1}$ by nudging the mixing parameters toward this probability distribution, so classes with higher losses receive a larger share of samples in the next time step. The scalar mixing rate, $\gamma$, is a user-defined step size hyperparameter that controls how aggressively $\alpha^{t-1}$ moves. We note that classical training is recovered by setting $\gamma = 0$.
 
Suppose that $\mathcal{H}$ is sufficiently expressive and can represent the true conditional expectation function, wherein there exists $\theta^* \in \mathbb{R}^m$ with $h_{\theta^*}(X) = \mathbb{E}[Y \mid X]$ almost surely. In the following proposition, we demonstrate that via gradient-based optimization under learn2mix training, the parameters converge to $\theta^*$, with the mixing proportions converging to a stable distribution that reflects the relative difficulty of each class under the optimal parameters.
\begin{proposition} \label{prop:mixing}
Let $\mathcal{L}(\theta^t), \mathcal{L}(\theta^*) \in \mathbb{R}^{k}$ denote the respective class-wise loss vectors for the model parameters at time $t$ and for the optimal model parameters. Suppose each class-wise loss $\mathcal{L}_i(\theta) \in \mathbb{R}$ is strongly convex in $\theta$, with strong convexity parameter $\mu_i \in \mathbb{R}_{> 0}$, $\forall i \in \{1,\ldots,k\}$, and each class-wise loss gradient $\smash{\nabla_{\theta} \mathcal{L}_i(\theta) \in \mathbb{R}^m}$ is Lipschitz continuous in $\theta$, having Lipschitz constant $L_i \in \mathbb{R}_{\geq 0}$, $\forall i \in \{1,\ldots,k\}$. Let $\smash{\mu^* = \min_{i \in \{1,\ldots,k \}} \mu_i}$, $\smash{L^* = \max_{i \in \{1,\ldots,k \}} L_i}$. Then, if the model parameters at time $t+1$ are obtained via the gradient of the loss for learn2mix training, where:
\begin{equation}
    \theta^{t+1} = \theta^{t} - \eta \nabla_{\theta} \mathcal{L}(\theta^{t}, \alpha^{t}), \quad \quad \text{with: \quad $\eta \in \mathbb{R}_{>0}$},
\end{equation}
It follows that for learning rate, $\eta \in (0, 2/L^*)$, and for mixing rate, $\gamma \in (0,1)$:
\begin{equation} \label{eq:mixing_update}
    \lim_{t \rightarrow \infty} \theta^t = \theta^*, \quad \quad \text{and:} \quad \lim_{t \rightarrow \infty} \alpha^t = \alpha^* = \frac{\mathcal{L}(\theta^*)}{\mathds{1}_k^T \mathcal{L}(\theta^*)}.
\end{equation}
\end{proposition}

The complete proof of Proposition \ref{prop:mixing} is provided in Section \ref{sec:proofs} of the Appendix. We now detail the convergence behavior of the learn2mix and classical training strategies, and suppose that $\alpha^{t-1} = \tilde{\alpha}$. We first present Corollary \ref{corr:loss_and_grad}, which will be used to prove the convergence result in Proposition \ref{prop:convergence}. This corollary leverages Lipschitz continuity and strong convexity to bound the loss gradient norm.
\begin{corollary} \label{corr:loss_and_grad}
Let $\mathcal{L}(\theta^t) \in \mathbb{R}^k$ denote the class-wise loss vector at time $t$. Suppose each class-wise loss, $\mathcal{L}_i(\theta) \in \mathbb{R}$, is strongly convex in $\theta$, with strong convexity parameter $\mu_i \in \mathbb{R}_{> 0}$, $\forall i \in \{1,\ldots,k\}$, and suppose each class-wise loss gradient $\smash{\nabla_{\theta} \mathcal{L}_i(\theta) \in \mathbb{R}^m}$ is Lipschitz continuous in $\theta$ with Lipschitz constant $L_i \in \mathbb{R}_{\geq 0}$, $\forall i \in \{1,\ldots,k\}$. Let $\smash{\mu^* = \min_{i \in \{1,\ldots,k \}} \mu_i}$, $\smash{L^* = \max_{i \in \{1,\ldots,k \}} L_i}$. Then, the following condition and inequality hold, $\forall \alpha \in [0,1]^k$ where $\smash{\sum_{i=1}^{\raisebox{-0.25ex}{$\scriptstyle k$}} \alpha_i = 1}$:
\begin{align}
    &\frac{\mu^*}{2} \| \theta^t - \theta^* \| \leq \| \nabla_{\theta} \mathcal{L}(\theta^t, \alpha) \| \leq L^* \| \theta^t - \theta^* \|, \\ \label{eq:grad_classical_learn2mix}
    &\text{Wherein: } \| \nabla_{\theta} \mathcal{L}(\theta^t, \alpha^t) \| + \| \nabla_{\theta} \mathcal{L}(\theta^t, \tilde{\alpha}) \| \leq 2L^* \| \theta^t - \theta^* \|.
\end{align}
\end{corollary}

The proof of Corollary \ref{corr:loss_and_grad} is provided in Section \ref{sec:proofs} of the Appendix --- we note that the inequality in Eq.~(\ref{eq:grad_classical_learn2mix}) relates the loss gradient norm under classical training with that under learn2mix training. We now present Proposition \ref{prop:convergence}, which demonstrates that under the condition expressed in Eq.~(\ref{eq:condition}), updates obtained via the gradient of the loss for learn2mix training bring the model parameters closer to the optimal solution than those obtained via the gradient of the loss for classical training.
\begin{proposition} \label{prop:convergence}
Let $\mathcal{L}(\theta^t), \mathcal{L}(\theta^*) \in \mathbb{R}^{k}$ denote the respective class-wise loss vectors for the model parameters at time $t$ and for the optimal model parameters. Suppose each class-wise loss, $\mathcal{L}_i(\theta) \in \mathbb{R}$ is strongly convex in $\theta$ with strong convexity parameter $\mu_i \in \mathbb{R}_{> 0}$, $\forall i \in \{1,\ldots,k\}$, and each class-wise loss gradient $\smash{\nabla_{\theta} \mathcal{L}_i(\theta) \in \mathbb{R}^m}$ is Lipschitz continuous in $\theta$, having Lipschitz constant $L_i \in \mathbb{R}_{\geq 0}$, $\forall i \in \{1,\ldots,k\}$. Moreover, suppose the loss gradient $\smash{\nabla_{\theta} \mathcal{L}(\theta, \alpha) \in \mathbb{R}^m}$ is Lipschitz continuous in $\alpha$, having Lipschitz constant $L_\alpha \in \mathbb{R}_{\geq 0}$, and let $\smash{\mu^* = \min_{i \in \{1,\ldots,k \}} \mu_i}$, $\smash{L^* = \max_{i \in \{1,\ldots,k \}} L_i}$. Then, if and only if the following condition holds:
\begin{align} \label{eq:condition}
    \Big[\Big(\frac{\mu^*}{2} - L^*\Big)\| \theta^t - \theta^* \|^2 + \tilde{\alpha}^T(\mathcal{L}(\theta^t) - \mathcal{L}(\theta^*)) \Big] \Big[\| \theta^t - \theta^* \| - (\mathcal{L}(\theta^t) - \mathcal{L}(\theta^*))\Big] > 0,
\end{align}
It follows that for every learning rate, $\eta > 0$, and for every mixing rate, $\gamma \in (0,\beta]$:
\begin{align} \label{eq:closer_step}
    \left\| \left(\theta^t - \eta \nabla_{\theta}\mathcal{L}(\theta^t, \alpha^t)\right) - \theta^{*} \right\| \leq \left\| \left(\theta^t - \eta \nabla_{\theta}\mathcal{L}(\theta^t, \tilde{\alpha})\right)  - \theta^{*} \right\|.
\end{align}
The complete formula for $\beta$ can be found in Section \ref{sec:proofs} of the Appendix.
\end{proposition}

The complete proof of Proposition \ref{prop:convergence} is provided in Section \ref{sec:proofs} of the Appendix.

\section{Algorithm} \label{sec:algorithm}
In this section, we outline our approach for training neural networks using learn2mix. The learn2mix mechanism comprises a bilevel optimization procedure, where we first update the neural network parameters, $\theta^t$, before updating the mixing parameters, $\alpha^t$, using the vector of class-wise losses, $\mathcal{L}(\theta^t)$. Considering the original training dataset, $J$, define $\smash{J_i = \{(x_j,y_j) \}_{j=1}^{\tilde{\alpha}_i N}}$, $\forall i \in \{1,\ldots,k \}$ as each class-specific training dataset, with $\smash{J = \bigcup_{i=1}^k J_i}$. These $k$ class-specific training datasets are leveraged to speed up batch formation under learn2mix. We consider neural network training using batched stochastic gradient descent, where for training epoch, $t$, the empirical loss is computed over $P = N/M$ total batches, where $M \in \mathbb{Z}^+$ is the batch size. Each batch is formed by sampling $\alpha_i^t M$ distinct examples from the $i$th class-specific training dataset, denoted as $S_i^p \subseteq J_i$, for $\smash{S^p = \biguplus_{i=1}^k S_i^p}$, where $\smash{\biguplus}$ is the set union operator that preserves duplicate elements. For learn2mix training, the class-wise errors, $\mathcal{L}_i(\theta^t), \forall i \in \{1,\ldots,k \}$, at training epoch $t$ are empirically computed as:
\begin{equation} \label{eq:learn2mix_classwise}
    \mathcal{L}_i(\theta^t) = \frac{1}{P} \sum_{p=1}^{P} \bigg[\frac{1}{\alpha_i^t M} \sum_{(x_j,y_j) \in S_i^p} \ell(h_{\theta^t}(x_j), y_j) \bigg],
\end{equation}
Where $\ell : \mathcal{Y} \times \mathcal{Y} \rightarrow \mathbb{R}_{\geq 0}$ is a bounded per-sample loss function and computes the error between the model prediction, $h_{\theta^t}(x_j)$, and the true label, $y_j$. Accordingly, the overall empirical loss at training epoch, $t$, under the learn2mix training mechanism is given by:
\begin{align} \label{eq:learn2mix_empirical}
   \mathcal{L}(\theta^t, \alpha^t) = \sum_{i=1}^{k} \alpha_i^t \mathcal{L}_i(\theta^t) = \sum_{i=1}^{k} \alpha_i^t \bigg[ \frac{1}{P} \sum_{p=1}^{P} \bigg[ \frac{1}{\alpha_i^t M} \sum_{(x_j,y_j) \in S_i^p} \ell(h_{\theta^t}(x_j), y_j) \bigg] \bigg].
\end{align}

\begin{algorithm}[t!]
\caption{Neural Network Training Via Learn2Mix}
\SetKwInput{KwInput}{Input}
\SetKwInput{KwOutput}{Output}
\label{alg:learn2mix_training}
\DontPrintSemicolon
\KwInput{$J$ (Original Training Dataset), $\theta$ (Initial NN Parameters), $\tilde{\alpha}$ (Initial Mixing Parameters), $\eta$ (Learning Rate), $\gamma$ (Mixing Rate), $M$ (Batch Size), $P$ (No. of Batches), $E$ (Epochs)}
\KwOutput{$\theta$ (Trained NN Parameters)}
\For{$i = 1,2,\ldots k$}{
    $J_i \gets \{(x_j, y_j)\}_{j=1}^{\alpha_i N}$ \ \ (Initialize class-specific training datasets) \\
    $\alpha_i \gets \tilde{\alpha}_i$ \ \ (Initialize time-varying mixing parameters)
}
\For{$epoch = 1,2,\ldots,E$}{  
    \For{$i = 1,2,\ldots,k$}{
        $J_i \gets \texttt{Shuffle}(J_i)$ \ \ (Randomly shuffle each class-specific training dataset)
    }
    \For{$p = 1,2,\ldots,P$}{
        \For{$i = 1,2,\ldots,k$}{
            $S_i^p \gets \texttt{Sample}(J_i, \alpha_i M)$ \ \ (Select $\alpha_i M$ distinct examples from $J_i$)}
        $\smash{S^p \gets \biguplus_{i=1}^{k} S_i^p}$ \ \ (Aggregate to form batch $S^p$) \\
        $\mathcal{L}^p(\theta, \alpha) \gets \frac{1}{M} \sum_{(x_j, y_j) \in S^p} \ell(h_{\theta}(x_j), y_j)$ \ \ (Compute loss on batch $S^p$)
    }
    $\mathcal{L}(\theta, \alpha) \gets \frac{1}{P} \sum_{p=1}^P \mathcal{L}^p(\theta, \alpha)$  \ \ (Obtain total loss) \\
    $\theta \gets \theta - \eta \nabla_{\theta} \mathcal{L}(\theta, \alpha)$ \ \ (Update model parameters, $\theta$) \\
    \For{$i = 1,2,\ldots,k$}{
        $\mathcal{L}_i(\theta) \gets \frac{1}{P} \sum_{p=1}^{P} \frac{1}{\alpha_i M} \sum_{(x_j, y_j) \in S_i^p} \ell(h_{\theta}(x_j), y_j)$ \ \ (Compute loss for class $i$)
    }
    $\alpha \gets \texttt{Update\_Mixing\_Params}(\alpha, \mathcal{L}(\theta), \gamma)$
}
\Return{$\theta$}
\end{algorithm}

Utilizing the empirical loss formulation from Eq.~(\ref{eq:learn2mix_empirical}), we now detail the algorithmic implementation of the learn2mix training methodology on a per-sample basis, for consistency with the mathematical preliminaries in Section \ref{sec:theoretical_results}. We note that the batch processing equivalent of this procedure is a trivial extension to the domain of matrices, and was used to generate the empirical results from Section \ref{sec:empirical_results}. Algorithm \ref{alg:learn2mix_training} outlines the primary training loop, where for each epoch, the class-specific datasets, $J_i$, are shuffled. Within each epoch, we iterate over the $P$ total batches, forming each batch by choosing $\alpha_i M$ examples from every $J_i$. The empirical loss within each batch is computed and aggregated to obtain the overall loss, $\mathcal{L}(\theta, \alpha)$, which is then used to update the neural network parameters through gradient descent. Lastly, the vector of class-wise losses, $\mathcal{L}(\theta)$, is calculated to inform the adjustment of the mixing parameters, $\alpha$, using Algorithm \ref{alg:update_mixing_parameters}.

Algorithm \ref{alg:update_mixing_parameters} outlines the method for adjusting class proportions using the mixing parameters, $\alpha$, based on the computed class-wise losses. For each class, $i \in \{1,\ldots,k \}$, we first calculate the normalized loss $L_i$ by dividing the class-specific loss $\mathcal{L}_i(\theta)$ by the total cumulative loss summed over all classes. Each mixing parameter, $\alpha_i$, is then updated incrementally towards this normalized loss value $L_i$. The magnitude of the update step is controlled by the mixing rate, $\gamma$, determining how quickly the proportions adapt. Thus, classes exhibiting higher relative losses are progressively given greater emphasis in subsequent training epochs, ensuring a balanced reduction of errors across all classes.

Finally, we recall that during the batch construction phase, for each class, $i \in \{1, \ldots, k\}$, we select $\alpha_i M$ examples from each $J_i$ to form the subset $S_i^p \subseteq J_i$. Given the dynamic nature of the mixing parameters, $\alpha$, it is possible that this cumulative selection across batches may exhaust all the samples within a particular $J_i$ before the epoch concludes. To address this, we incorporate a cyclic selection mechanism. Formally, we define an index $\tau_i^p$, $\forall i \in \{1, \ldots, k\}$ and $p \in \{1, \ldots, P \}$, such that:
\begin{equation}
    \tau_i^p = \left( \tau_i^{p-1} + \alpha_i M \right) \mod \tilde{\alpha}_i N,
\end{equation}

Where $\tau_i^0 = 0$, $\forall i \in \{1,\ldots,k \}$. Accordingly, when selecting $S_i^p$, if $\tau_i^{p-1} + \alpha_i M > \tilde{\alpha}_i N$, we wrap around to the beginning of $J_i$, effectively resetting the selection index, $\tau_i^p$ --- this ensures that every example in $J_i$ is selected uniformly and repeatedly as needed throughout the training process. Thus, the selection procedure to construct $S_i^p$ is defined as:
\begin{equation}
    S_i^p = \biguplus\nolimits_{w=0}^{\alpha_i M - 1} J_i\left[ (\tau_i^{p-1} + w) \mod \tilde{\alpha}_i N \right].
\end{equation}

This cyclic selection procedure ensures that the required number of samples, $\alpha_i M$, for each class in every batch is maintained, even as $\alpha_i$ is adaptively updated across epochs.

\begin{algorithm}[t!]
\caption{Updating Mixing Parameters Via Learn2Mix}
\SetKwInput{KwInput}{Input}
\SetKwInput{KwOutput}{Output}
\label{alg:update_mixing_parameters}
\DontPrintSemicolon
\KwInput{$\alpha$ (Previous Mixing Parameters), $\mathcal{L}(\theta)$ (Class-wise loss vector), $\gamma$ (Mixing Rate)}
\KwOutput{$\alpha$ (Updated Mixing Parameters)}
\For{$i = 1,2,\ldots,k$}{
        $L_i \gets \frac{\mathcal{L}_i(\theta)}{\sum_{j=1}^{k} \mathcal{L}_j(\theta)}$ \ \ (Compute normalized losses) \\
        $\alpha_i \gets \alpha_i + \gamma \left(L_i - \alpha_i \right)$ \ \ (Update mixing parameters)
}
\Return{$\alpha$}
\end{algorithm}

\section{Empirical Results} \label{sec:empirical_results}
In this section, we present our empirical results on classification, regression, and image reconstruction tasks, across both benchmark and modified class imbalanced datasets. We first present the classification results on three benchmark datasets (MNIST \citep{deng2012mnist}, Fashion-MNIST \citep{xiao2017fashion}, CIFAR-10 \citep{krizhevsky2009learning}), and three standard datasets with manually imbalanced classes (Imagenette \citep{imagenette}, CIFAR-100 \citep{krizhevsky2009learning}, and IMDB \citep{maas2011imdb}). We note that for the imbalanced case, we only introduce the manual class-imbalancing to the training dataset, $J$, where the test dataset, $\smash{K = \{ (x_j,y_j)\}_{j=1}^{N_{\text{test}}}}$, is not changed. This choice ensures that the generalization performance of the network is benchmarked in a class-balanced setting. Next, for the regression task, we study two benchmark datasets with manually imbalanced classes (Wine Quality \citep{cortez2009modeling}, and California Housing \citep{geron2022hands}), and a synthetic mean estimation task, where the manual class-imbalancing parallels that of the classification case. Finally, we reconsider the MNIST, Fashion MNIST and CIFAR-10 datasets for image reconstruction, again with manual class-imbalancing. The comprehensive description of these datasets and class-imbalancing strategies is in Section \ref{sec:dataset_descriptions} of the Appendix. For further performance verification, we include ablation studies on architecture, optimizer, batch size, learning rate, and worst-class error in Section \ref{sec:additional_results} of the Appendix.

The intuition behind the application of learn2mix to regression and reconstruction tasks stems from its ability to adaptively handle different data distributions. For regression tasks with a categorical variable taking $k$ distinct values, the samples from $J$ that correspond to each of the $k$ values, can be aggregated to obtain each class-specific training dataset, $J_i$. Here, each $J_i$ denotes a distinct underlying data distribution. As in the classification case, learn2mix will adaptively adjust the class-specific dataset proportions during training. For image reconstruction, we can similarly treat the $k$ distinct classes being reconstructed as the values taken by a categorical variable, paralleling the regression context. This formulation supports the adaptive adjustment of class proportions under learn2mix training.

For the evaluations that follow, all training was performed on an NVIDIA GEForce RTX 3090 GPU. To ensure a fair comparison between learn2mix and classical training, we utilize the same learning rate, $\eta$, and neural network architecture with initialized parameters, $\theta$, across all experiments for a given dataset. Additionally, we train each neural network through learn2mix (with mixing rate $\gamma$) and classical training for $\smash{\underline{E}}$ seconds (or $E$ epochs), where $\smash{\underline{E}}$ is dataset dependent \footnote{Practically, we observe that choosing $\gamma \in [0.01,0.1]$ improves performance (see Section \ref{sec:additional_results} of the Appendix).}. In classification tasks, we also benchmark learn2mix and classical training versus `FCL training', `SMOTE training', `IS training', and `CURR training' (training using focal loss \citep{lin2017focal}, SMOTE oversampling \citep{chawla2002smote}, importance sampling \citep{katharopoulos2018not}, and curriculum learning \citep{hacohen2019power} --- see Sections \ref{sec:focal_training}, \ref{sec:smote_training},  \ref{sec:is_training}, and \ref{sec:curr_training} of the Appendix). The complete list of model architectures and hyperparameters is in Section \ref{sec:experiment_details} of the Appendix.

\begin{table*}[t!]
\caption{Test accuracies for learn2mix (L2M), classical (CL), FCL, SMOTE, IS, CURR training.}
\label{tab:classification_performance}
\centering
\footnotesize  
\begin{tabular}{l|c|c|c|c|c|c}
\noalign{\hrule height 1.2pt}
\multicolumn{7}{c}{\textbf{\texttt{Elapsed Time}}: $\boldsymbol{\underline{t} = 0.25\underline{E}}$ s}  \\
\hline
\textbf{Dataset} & \textbf{MNIST} & \textbf{Fsh. MNIST} & \textbf{CIFAR-10} & \textbf{Imagenette} & \textbf{CIFAR-100} & \textbf{IMDB} \\
\hline
\textbf{Acc (L2M)} & \textcolor{Red}{$95.42 \scriptstyle \pm 0.28$} & \textcolor{Red}{$77.62 \scriptstyle \pm 0.69$} & \textcolor{Red}{$51.34 \scriptstyle \pm 0.13$} & \textcolor{Red}{$24.12 \scriptstyle \pm 0.46$} & \textcolor{Red}{$30.03 \scriptstyle \pm 1.30$} & \textcolor{Red}{$70.28 \scriptstyle \pm 1.66$} \\
\textbf{Acc (CL)}  & $93.14 \scriptstyle \pm 0.47$ & $74.13 \scriptstyle \pm 0.73$ & $49.26 \scriptstyle \pm 0.15$ & $15.55 \scriptstyle \pm 0.13$ & $23.23 \scriptstyle \pm 1.99$ & $50.13 \scriptstyle \pm 0.15$ \\
\textbf{Acc (FCL)} & $91.32 \scriptstyle \pm 0.57$ & $74.08 \scriptstyle \pm 0.75$ & $47.90 \scriptstyle \pm 0.22$ & $20.11 \scriptstyle \pm 0.37$ & $27.15 \scriptstyle \pm 1.13$ & $50.30 \scriptstyle \pm 0.67$ \\
\textbf{Acc (SMOTE)} & $92.41 \scriptstyle \pm 0.71$ & $73.67 \scriptstyle \pm 0.61$ & $47.76 \scriptstyle \pm 0.15$ & $23.19 \scriptstyle \pm 0.46$ & $23.93 \scriptstyle \pm 2.35$ & $50.94 \scriptstyle \pm 0.05$ \\
\textbf{Acc (IS)} & $92.44 \scriptstyle \pm 0.63$ & $74.23 \scriptstyle \pm 0.29$ & $47.40 \scriptstyle \pm 0.51$ & $23.10 \scriptstyle \pm 0.39$ & $27.97 \scriptstyle \pm 0.67$ & $58.48 \scriptstyle \pm 0.64$ \\
\textbf{Acc (CURR)} & $93.30 \scriptstyle \pm 0.54$ & $75.06 \scriptstyle \pm 0.63$ & $49.11 \scriptstyle \pm 0.25$ & $18.82 \scriptstyle \pm 0.37$ & $27.15 \scriptstyle \pm 0.10$ & $50.02 \scriptstyle \pm 0.04$ \\
\hline
\multicolumn{7}{c}{\textbf{\texttt{Elapsed Time}}: $\boldsymbol{\underline{t} = 0.5\underline{E}}$ s} \\
\hline
\textbf{Dataset} & \textbf{MNIST} & \textbf{Fsh. MNIST} & \textbf{CIFAR-10} & \textbf{Imagenette} & \textbf{CIFAR-100} & \textbf{IMDB} \\
\hline
\textbf{Acc (L2M)} & \textcolor{Red}{$97.61 \scriptstyle \pm 0.15$} & \textcolor{Red}{$83.16 \scriptstyle \pm 0.87$} & \textcolor{Red}{$55.84 \scriptstyle \pm 0.19$} & \textcolor{Red}{$33.64 \scriptstyle \pm 0.42$} & \textcolor{Red}{$46.80 \scriptstyle \pm 0.54$} & \textcolor{Red}{$76.02 \scriptstyle \pm 2.77$} \\
\textbf{Acc (CL)}  & $96.74 \scriptstyle \pm 0.10$ & $79.75 \scriptstyle \pm 0.83$ & $54.50 \scriptstyle \pm 0.34$ & $23.63 \scriptstyle \pm 0.33$ & $43.00 \scriptstyle \pm 0.73$ & $74.99 \scriptstyle \pm 0.57$ \\
\textbf{Acc (FCL)} & $95.92 \scriptstyle \pm 0.13$ & $78.94 \scriptstyle \pm 0.82$ & $54.15 \scriptstyle \pm 0.06$ & $29.44 \scriptstyle \pm 0.43$ & $40.26 \scriptstyle \pm 0.55$ & $68.30 \scriptstyle \pm 3.21$ \\
\textbf{Acc (SMOTE)} & $96.51 \scriptstyle \pm 0.16$ & $79.17 \scriptstyle \pm 0.50$ & $53.72 \scriptstyle \pm 0.16$ & $28.90 \scriptstyle \pm 0.43$ & $39.10 \scriptstyle \pm 1.63$ & $62.72 \scriptstyle \pm 0.54$ \\
\textbf{Acc (IS)} & $96.60 \scriptstyle \pm 0.25$ & $79.65 \scriptstyle \pm 0.38$ & $52.56 \scriptstyle \pm 0.39$ & $28.52 \scriptstyle \pm 0.32$ & $42.61 \scriptstyle \pm 2.61$ & $74.00 \scriptstyle \pm 0.81$ \\
\textbf{Acc (CURR)} & $96.53 \scriptstyle \pm 0.16$ & $79.08 \scriptstyle \pm 0.58$ & $53.49 \scriptstyle \pm 0.33$ & $27.26 \scriptstyle \pm 0.79$ & $39.48 \scriptstyle \pm 2.22$ & $71.68 \scriptstyle \pm 0.55$ \\
\hline
\multicolumn{7}{c}{\textbf{\texttt{Elapsed Time}}: $\boldsymbol{\underline{t} = \underline{E}}$ s} \\
\hline
\textbf{Dataset} & \textbf{MNIST} & \textbf{Fsh. MNIST} & \textbf{CIFAR-10} & \textbf{Imagenette} & \textbf{CIFAR-100} & \textbf{IMDB} \\
\hline
\textbf{Acc (L2M)} & \textcolor{Red}{$98.46 \scriptstyle \pm 0.14$} & \textcolor{Red}{$85.85 \scriptstyle \pm 0.47$} & \textcolor{Red}{$60.49 \scriptstyle \pm 0.26$} & \textcolor{Red}{$42.95 \scriptstyle \pm 0.33$} & \textcolor{Red}{$54.50 \scriptstyle \pm 0.73$} & \textcolor{Red}{$82.38 \scriptstyle \pm 0.59$} \\
\textbf{Acc (CL)}  & $98.14 \scriptstyle \pm 0.14$ & $84.23 \scriptstyle \pm 0.60$ & $59.62 \scriptstyle \pm 0.16$ & $34.53 \scriptstyle \pm 0.33$ & $52.30 \scriptstyle \pm 0.36$ & $80.84 \scriptstyle \pm 0.71$ \\
\textbf{Acc (FCL)} & $97.86 \scriptstyle \pm 0.08$ & $83.68 \scriptstyle \pm 0.61$ & $59.37 \scriptstyle \pm 0.64$ & $40.60 \scriptstyle \pm 0.71$ & $49.33 \scriptstyle \pm 0.97$ & $79.09 \scriptstyle \pm 2.58$ \\
\textbf{Acc (SMOTE)} & $98.09 \scriptstyle \pm 0.07$ & $83.57 \scriptstyle \pm 1.06$ & $58.46 \scriptstyle \pm 0.15$ & $39.59 \scriptstyle \pm 0.29$ & $50.63 \scriptstyle \pm 1.02$ & $74.64 \scriptstyle \pm 1.28$ \\
\textbf{Acc (IS)} & $98.14 \scriptstyle \pm 0.14$ & $84.33 \scriptstyle \pm 0.29$ & $57.44 \scriptstyle \pm 0.42$ & $35.33 \scriptstyle \pm 0.43$ & $52.83 \scriptstyle \pm 0.34$ & $79.08 \scriptstyle \pm 0.57$ \\
\textbf{Acc (CURR)} & $98.13 \scriptstyle \pm 0.05$ & $83.32 \scriptstyle \pm 0.43$ & $59.15 \scriptstyle \pm 0.42$ & $35.26 \scriptstyle \pm 0.48$ & $50.88 \scriptstyle \pm 0.79$ & $80.04 \scriptstyle \pm 0.25$ \\
\noalign{\hrule height 1.2pt}
\end{tabular}
\end{table*}

\begin{figure*}[t!]
    \centering
    \begin{subfigure}{0.32\textwidth}
        \includegraphics[width=\textwidth]{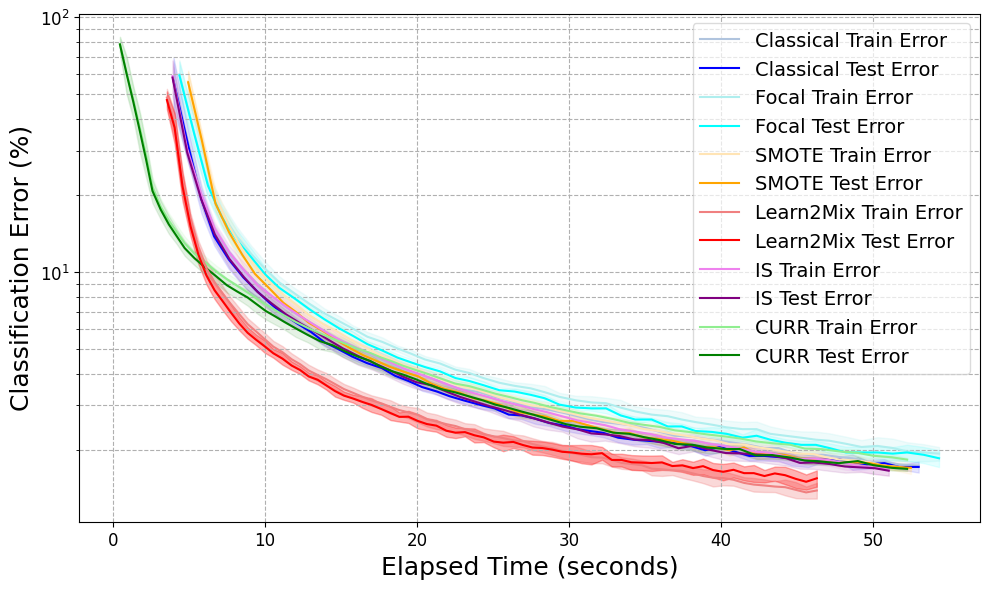}
        \caption{MNIST}
    \end{subfigure}
        \begin{subfigure}{0.32\textwidth}
        \includegraphics[width=\textwidth]{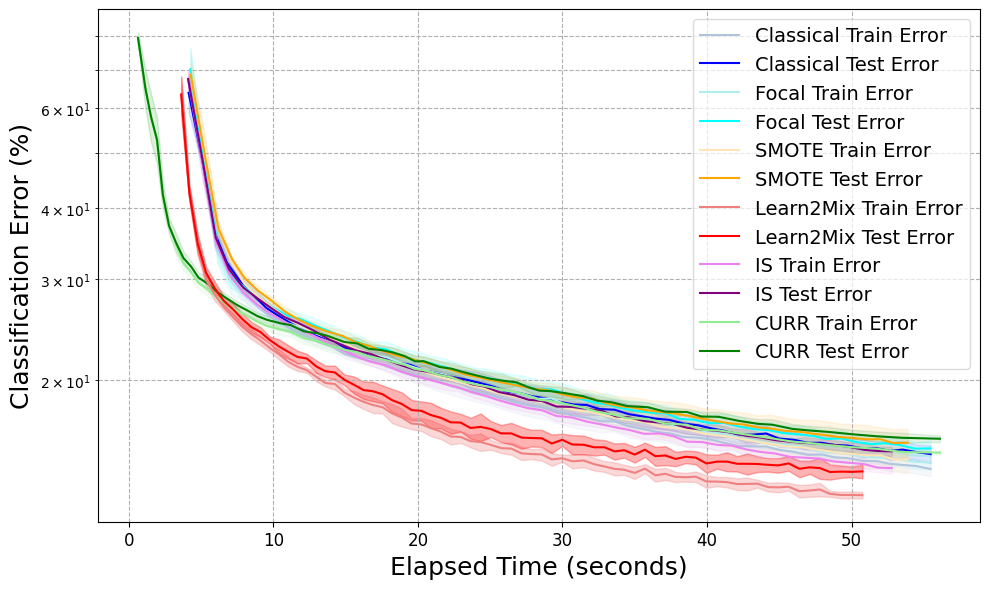}
        \caption{Fashion MNIST}
    \end{subfigure}
    \begin{subfigure}{0.32\textwidth}
        \includegraphics[width=\textwidth]{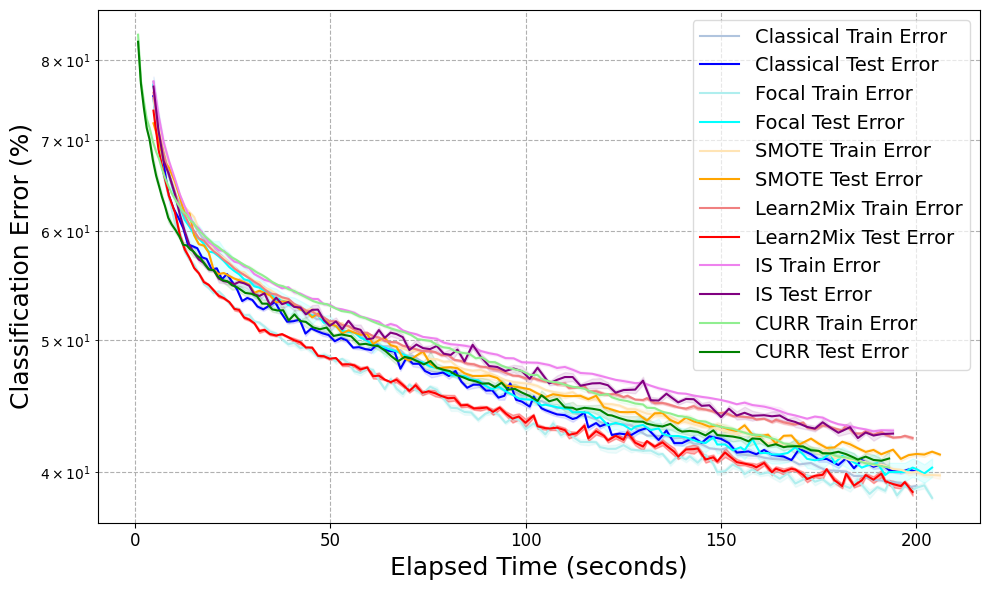}
        \caption{CIFAR-10}
    \end{subfigure}
\\
        \begin{subfigure}{0.32\textwidth}
        \includegraphics[width=\textwidth]{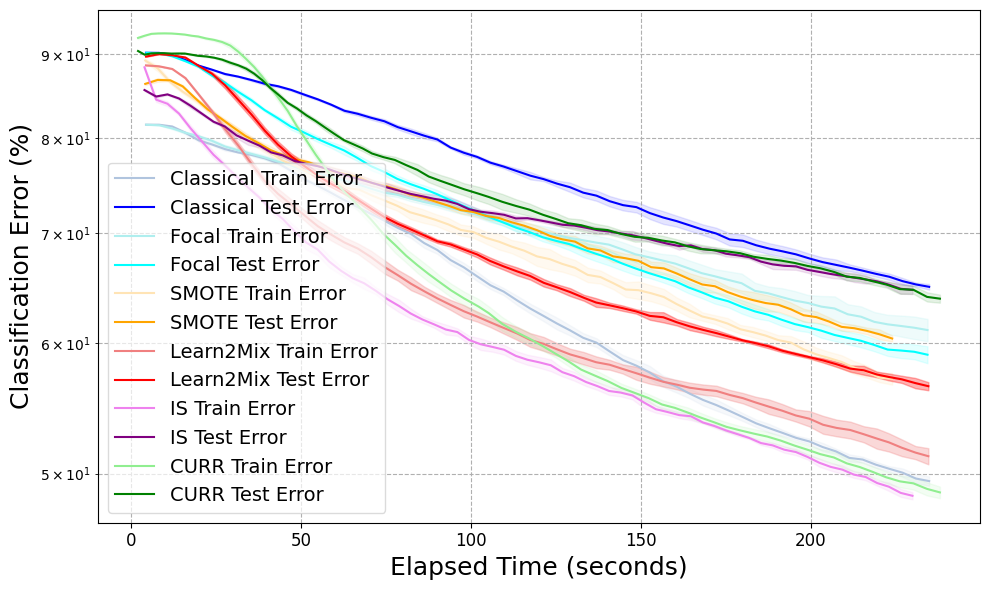}
        \caption{Imagenette}
    \end{subfigure}
    \begin{subfigure}{0.32\textwidth}
        \includegraphics[width=\textwidth]{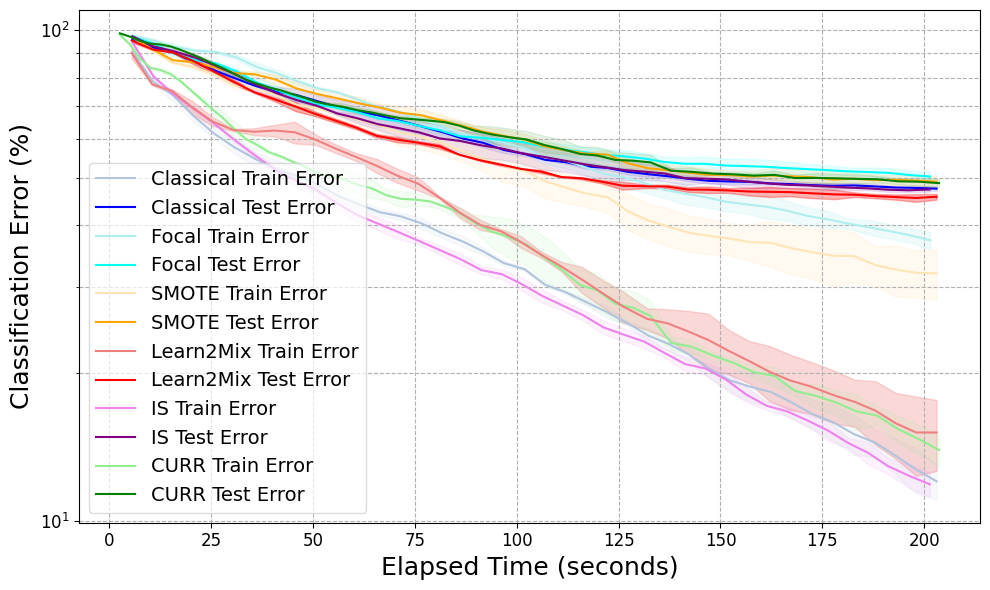}
        \caption{CIFAR-100}
    \end{subfigure}
    \begin{subfigure}{0.32\textwidth}
        \includegraphics[width=\textwidth]{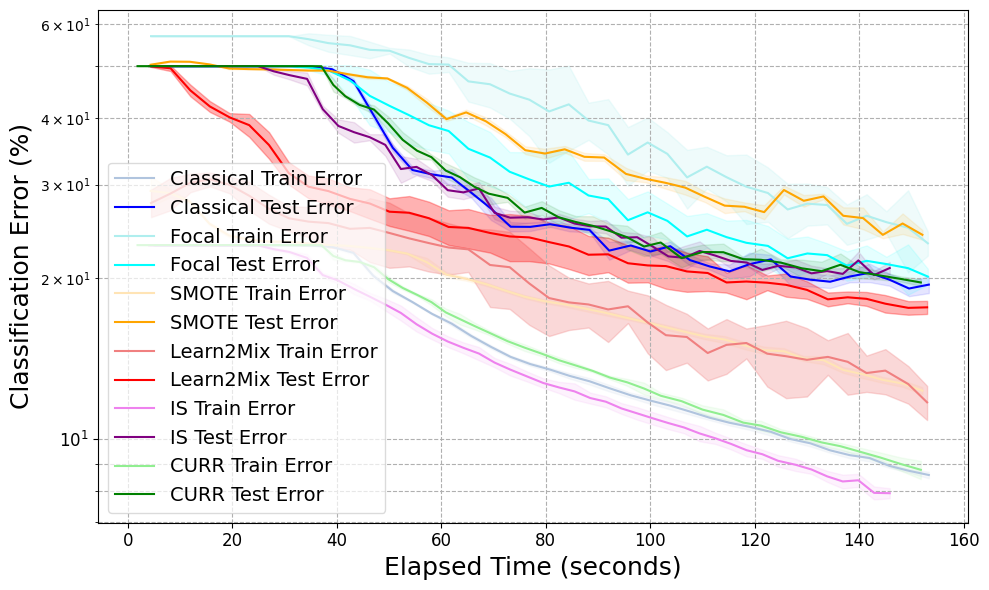}
        \caption{IMDB}
    \end{subfigure} 
    \caption{Comparing model classification errors for learn2mix, classical, FCL, SMOTE, IS, and CURR training. The x-axis is the elapsed [training] time, while the y-axis is the classification error.} \label{fig:classification_performance}
\end{figure*}

\subsection{Classification Tasks} \label{sec:classification_tasks}
As illustrated in Table \ref{tab:classification_performance} and Figure \ref{fig:classification_performance}, we observe a consistent trend across all considered classification benchmarks, whereby neural networks trained using learn2mix converge faster than their classically-trained, FCL-trained, SMOTE-trained, IS-trained, and CURR-trained counterparts. We first consider MNIST, and train LeNet-5 \citep{lecun1998gradient} via the Adam optimizer \citep{kingma2014adam} and Cross Entropy Loss for $\underline{E} = 50$ s, leveraging learn2mix, classical, FCL, SMOTE, IS, and CURR training. We see that the learn2mix-trained CNN converges faster, eclipsing a test accuracy of $98\%$ within $30$ s, whereas the remaining CNNs achieve this test accuracy after $40$ s. We next consider the more challenging Fashion MNIST dataset, and train Large LeNet-5 for $\underline{E} = 50$ s with the Adam optimizer and Cross Entropy Loss, leveraging learn2mix, classical, FCL, SMOTE, IS, and CURR training. Paralleling MNIST, we observe that the learn2mix-trained CNN converges faster, yielding a test accuracy of $83\%$ within $20$ s, whereas the other CNNs achieve this test accuracy after $33$ s. The last class-balanced benchmark dataset we investigate is CIFAR-10, which offers a greater challenge than MNIST and Fashion MNIST. We train Large LeNet for $\underline{E} = 200$ s using the Adam optimizer and Cross Entropy Loss, utilizing learn2mix, classical, FCL, SMOTE, IS, and CURR training. We observe that the learn2mix-trained CNN achieves faster convergence, yielding a test accuracy of $60\%$ after $170$ s, whereas the remaining CNNs exceed this test accuracy after $200$ s. Cumulatively, these evaluations demonstrate the efficacy of learn2mix training in settings with balanced classes, wherein the adaptive adjustment of class proportions accelerates convergence.

We now consider several class-imbalanced training datasets. We first benchmark Imagenette, which comprises a subset of 10 classes from ImageNet \citep{deng2009imagenet}, and modify the training dataset so the number of samples from each class, $i \in \{1,\ldots,k\}$, in $J$ decreases linearly. We train ResNet-18 \citep{he2016deep} with the Adam optimizer and Cross Entropy Loss for $\underline{E} = 230$ s, using learn2mix, classical, FCL, SMOTE, IS, and CURR training. We see the learn2mix-trained ResNet-18 converges faster, achieving a test accuracy of $40\%$ after $185$ s, whereas only the FCL-trained model achieves this test accuracy after $230$ s. We now consider CIFAR-100, and modify the training dataset so the number of samples from each class, $i \in \{1,\ldots,k\}$, in $J$ decreases logarithmically. We train MobileNet-V3 Small \citep{howard2019searching} for $\underline{E} = 200$ s leveraging the Adam optimizer and Cross Entropy Loss, using learn2mix, classical, FCL, SMOTE, IS, and CURR training. We see that the learn2mix-trained MobileNet-V3 Small model converges faster, achieving a test accuracy of $50\%$ within $120$ s, whereas the other models exceed this test accuracy after $140$ s. As the $k=100$ mixing parameters are a small fraction of the total model parameters, this overhead is negligible. For IMDB, we modify the training dataset so the positive class keeps $30\%$ of its original samples. We train a transformer for $\underline{E} = 150$ s with the Adam optimizer and Cross Entropy Loss, using learn2mix, classical, FCL, SMOTE, IS, and CURR training. We see the learn2mix-trained transformer converges faster, reaching a test accuracy of $70\%$ within $35$ s, whereas the other models exceed this test accuracy after $60$ s. 

In the above evaluations, we see learn2mix not only accelerates convergence, but also has a tighter alignment between training and test errors versus classical training. This correspondence indicates reduced overfitting, as learn2mix inherently adjusts class proportions based on class-specific error rates, $L_i$. By biasing the optimization procedure away from the original class distribution and towards $L_i$, learn2mix achieves improved generalization. We note this property is not unique to classification and also applies to regression and reconstruction (see Sections \ref{sec:regression_tasks} and \ref{sec:reconstruction_tasks}).

\subsection{Regression Tasks} \label{sec:regression_tasks}
As illustrated in Table \ref{tab:regression_performance} and Figure \ref{fig:regression_performance}, we observe that learn2mix maintains accelerated convergence in the regression context, wherein all the considered datasets are class imbalanced. We first consider the synthetic Mean Estimation dataset, which comprises sets of samples gathered from $k = 4$ unique distributions and their associated means. Using the Adam optimizer and Mean Squared Error (MSE) Loss, we train a fully connected network for $E = 500$ epochs on Mean Estimation via learn2mix and classical training. We see that the learn2mix-trained neural network converges rapidly, achieving a test error below $2.0$ after 100 epochs, at which point the classically-trained network has a test error of $13.0$. For the Wine Quality dataset, we modify the training dataset so the white wine class has $10\%$ of its original samples. Using the Adam optimizer and MSE Loss, we train a fully connected network for $E = 300$ epochs on Wine Quality via learn2mix and classical training. We observe that the learn2mix-trained neural network yields faster convergence, achieving a test error below $2.5$ after 200 epochs, at which point the classically-trained network has a test error of $5.0$. Finally, on the California Housing dataset, we modify the training dataset such that three of the classes have $5\%$ of their original samples. Using the Adam optimizer and MSE Loss, we train a fully connected network for $E = 1200$ epochs on California Housing via learn2mix and classical training. We again see that the learn2mix-trained network converges faster, achieving a test error below $0.8$ after 1200 epochs, while the classically-trained network has a test error of $0.99$. These empirical evaluations support our previous intuition pertaining to the extension of learn2mix to imbalanced regression settings. 

\subsection{Image Reconstruction Tasks} \label{sec:reconstruction_tasks}
\begin{figure*}[t!]
    \centering
    \begin{subfigure}{0.313\textwidth}
        \includegraphics[width=\textwidth]{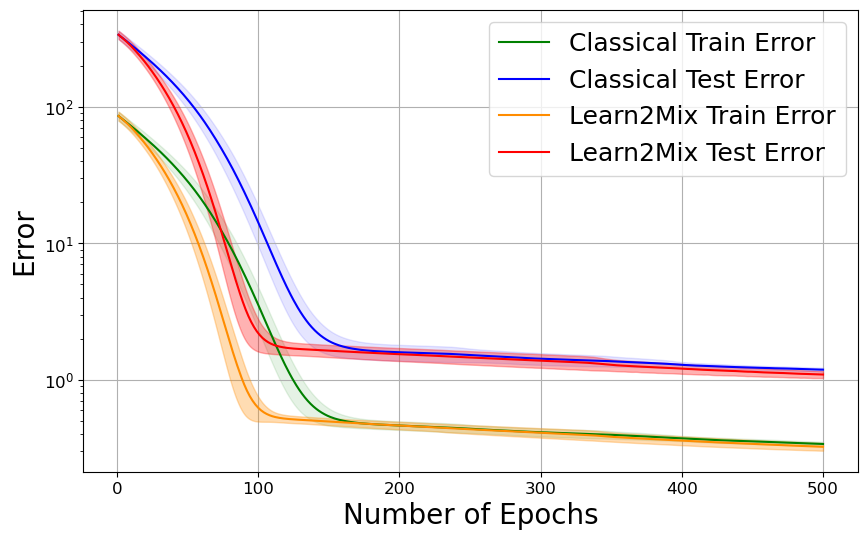}
        \caption{Mean Estimation}
    \end{subfigure}
    \begin{subfigure}{0.313\textwidth}
        \includegraphics[width=\textwidth]{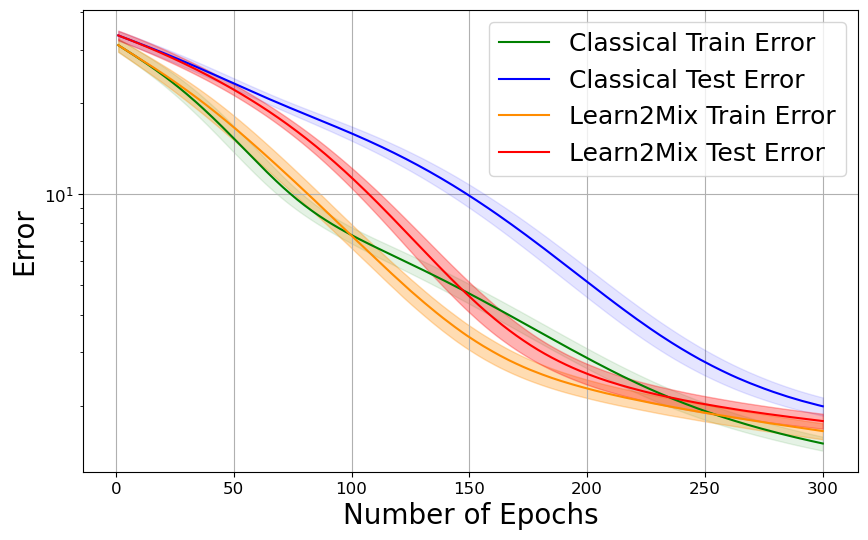}
        \caption{Wine Quality}
    \end{subfigure}
    \begin{subfigure}{0.313\textwidth}
        \includegraphics[width=\textwidth]{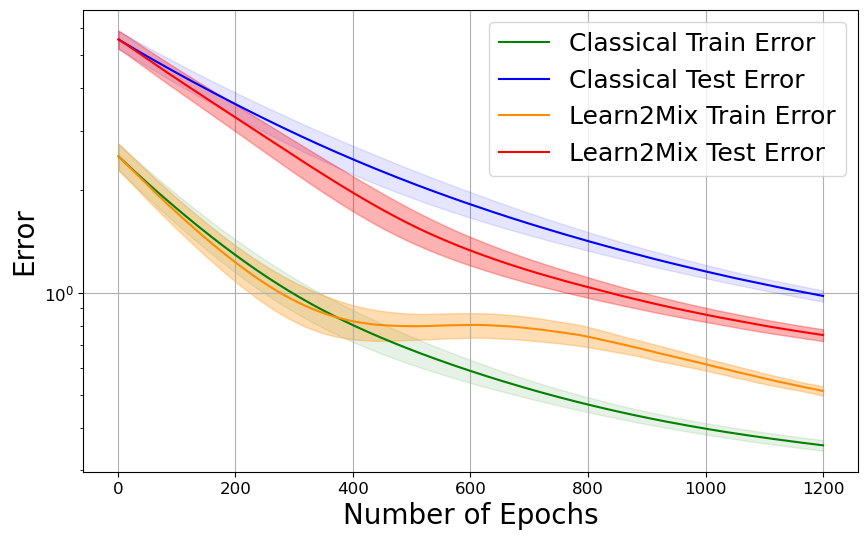}
        \caption{California Housing}
    \end{subfigure}
\\
    \begin{subfigure}{0.315\textwidth}
        \includegraphics[width=\textwidth]{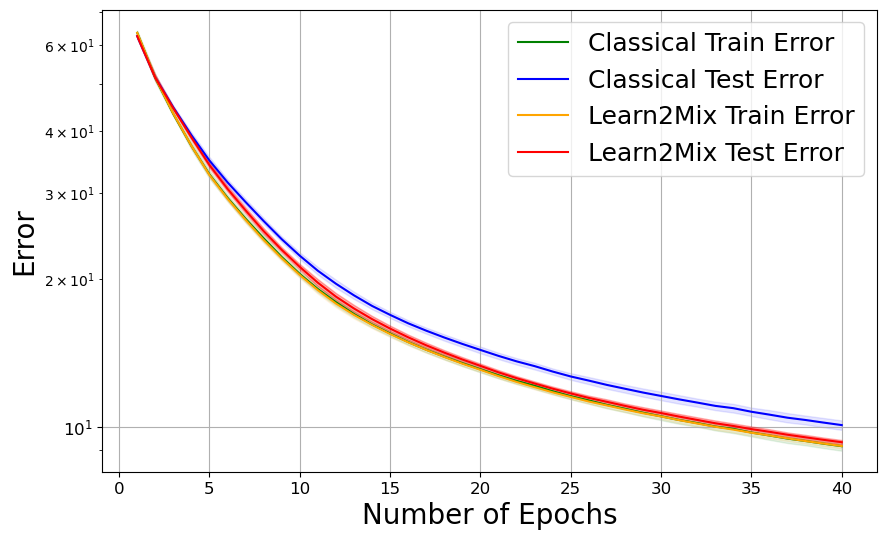}
        \caption{MNIST}
    \end{subfigure}
    \begin{subfigure}{0.315\textwidth}
        \includegraphics[width=\textwidth]{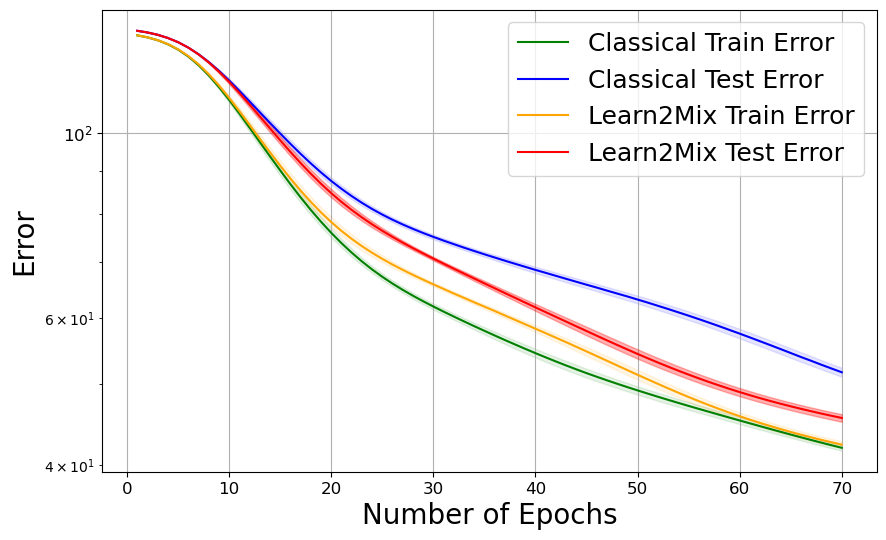}
        \caption{Fashion MNIST}
    \end{subfigure}
    \begin{subfigure}{0.315\textwidth}
        \includegraphics[width=\textwidth]{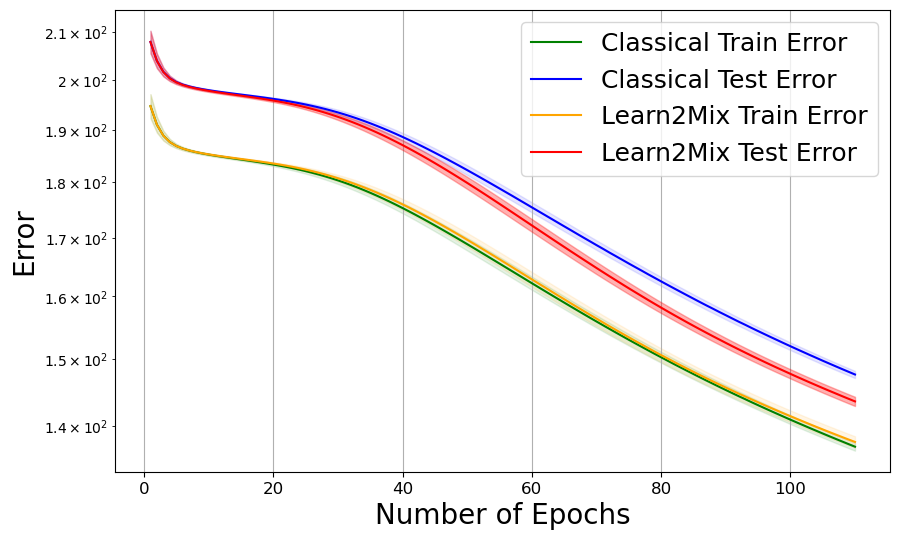}
        \caption{CIFAR-10}
    \end{subfigure}
    \caption{Comparing model performance errors for classical training and learn2mix training. The x-axis is the number of elapsed training epochs, while the y-axis is the mean squared error (MSE).} \label{fig:regression_performance}
\end{figure*}

Per Table \ref{tab:regression_performance} and Figure \ref{fig:regression_performance}, we note that the class-imbalanced image reconstruction tasks also observe faster convergence using learn2mix. For the MNIST case, we modify the training dataset such that half of the classes retain $20\%$ of their original samples. Leveraging the Adam optimizer and MSE Loss, we train an autoencoder for $E = 40$ epochs on MNIST using learn2mix and classical training. We observe that the learn2mix-trained autoencoder exhibits improved convergence, achieving a test error less than $1.0$ after 35 epochs, which the classically-trained autoencoder achieves after 40 epochs. Correspondingly, for Fashion MNIST, we modify the training dataset such that half of the classes retain $20\%$ of their original samples (paralleling MNIST). Using the Adam optimizer and MSE Loss, we train an autoencoder for $E = 70$ epochs on Fashion MNIST, leveraging learn2mix and classical training. We observe that the learn2mix-trained autoencoder converges faster, achieving a test error below $54.0$ after 50 epochs, which the classically-trained autoencoder achieves after 65 epochs. We also consider CIFAR-10, wherein we modify the training dataset such that all but two classes retain $20\%$ of their original samples. Utilizing the Adam optimizer and MSE Loss, we train an autoencoder for $E = 110$ epochs on CIFAR-10, leveraging learn2mix and classical training. We observe that the learn2mix-trained autoencoder also converges faster and achieves a test error below $148.0$ after 100 epochs, which the classically-trained autoencoder achieves after 110 epochs. 

\begin{table*}[t!]
\caption{Test mean squared error for learn2mix (L2M) and classical (CL) training.}
\label{tab:regression_performance}
\begin{center}
\begin{tabular}{l|cc|cc|cc}
\noalign{\hrule height 1.2pt}
& \multicolumn{2}{c|}{\textbf{\texttt{Epoch}} $\boldsymbol{t = 0.25E}$} & \multicolumn{2}{c|}{\textbf{\texttt{Epoch}} $\boldsymbol{t = 0.5E}$} & \multicolumn{2}{c}{\textbf{\texttt{Epoch}} $\boldsymbol{t = E}$} \\
\textbf{Dataset} & \textbf{Err (L2M)} & \textbf{Err (CL)} & \textbf{Err (L2M)} & \textbf{Err (CL)} & \textbf{Err (L2M)} & \textbf{Err (CL)} \\
\hline
\textbf{Mean Estim.} & \textcolor{Red}{$1.81 \scriptstyle \pm 0.84$} & $6.51 \scriptstyle \pm 1.52$ & \textcolor{Red}{$1.45 \scriptstyle \pm 0.26$} & $1.52 \scriptstyle \pm 0.27$ & \textcolor{Red}{$1.07 \scriptstyle \pm 0.09$} & $1.17 \scriptstyle \pm 0.06$ \\
\textbf{Wine Quality} & \textcolor{Red}{$17.7 \scriptstyle \pm 1.64$} & $19.8 \scriptstyle \pm 1.51$ & \textcolor{Red}{$4.26 \scriptstyle \pm 1.55$} & $9.72 \scriptstyle \pm 1.94$ & \textcolor{Red}{$1.75 \scriptstyle \pm 0.21$} & $2.03 \scriptstyle \pm 0.18$ \\
\textbf{Cali. Housing} & \textcolor{Red}{$2.52 \scriptstyle \pm 0.68$} & $2.95 \scriptstyle \pm 0.67$ & \textcolor{Red}{$1.33 \scriptstyle \pm 0.32$} & $1.82 \scriptstyle \pm 0.39$ & \textcolor{Red}{$0.77 \scriptstyle \pm 0.08$} & $0.99 \scriptstyle \pm 0.10$ \\
\textbf{MNIST} & \textcolor{Red}{$19.6 \scriptstyle \pm 0.81$} & $20.8 \scriptstyle \pm 0.93$ & \textcolor{Red}{$12.9 \scriptstyle \pm 0.39$} & $14.0 \scriptstyle \pm 0.52$ & \textcolor{Red}{$9.31 \scriptstyle \pm 0.24$} & $10.1 \scriptstyle \pm 0.56$ \\
\textbf{Fsh. MNIST} & \textcolor{Red}{$89.3 \scriptstyle \pm 2.63$} & $91.9 \scriptstyle \pm 2.37$ & \textcolor{Red}{$65.1 \scriptstyle \pm 1.21$} & $70.9 \scriptstyle \pm 1.28$ & \textcolor{Red}{$45.5 \scriptstyle \pm 1.21$} & $51.6 \scriptstyle \pm 1.60$ \\
\textbf{CIFAR-10} & \textcolor{Red}{$193 \scriptstyle \pm 1.23$} & $194 \scriptstyle \pm 1.98$ & \textcolor{Red}{$175 \scriptstyle \pm 2.85$} & $179 \scriptstyle \pm 3.87$ & \textcolor{Red}{$144 \scriptstyle \pm 1.71$} & $148 \scriptstyle \pm 1.37$ \\
\noalign{\hrule height 1.2pt}
\end{tabular}
\end{center}
\end{table*}

\section{Conclusion} \label{sec:conclusion}
In this work, we presented \textit{learn2mix}, a new training strategy that adaptively modifies class proportions in batches via real-time class-wise error rates, accelerating model convergence. We formalized the learn2mix mechanism through a bilevel optimization framework, and outlined its theoretical advantages in aligning class proportions with optimal error rates. Empirical evaluations across classification, regression, and reconstruction tasks on both balanced and imbalanced datasets confirmed that learn2mix not only accelerates convergence compared to classical training methods, but also reduces overfitting in the presence of class-imbalances. Accordingly, models trained with learn2mix achieved improved performance in constrained training regimes. Our findings underscore the potential of dynamic batch composition strategies in optimizing neural network training, paving the way for more efficient and robust machine learning models in resource-constrained environments.

\ack{Shyam Venkatasubramanian and Vahid Tarokh were supported in part by the Air Force Office of Scientific Research under award FA9550-21-1-0235.}

\bibliography{neurips_2025.bib}
\bibliographystyle{plainnat}


\newpage
\appendix
{\Huge \textbf{Appendix}}

\section{Proofs of the Theoretical Results} \label{sec:proofs}
In this section, we present the proofs of the theoretical results outlined in the main text.

\begin{proposition1}
Let $\mathcal{L}(\theta^t), \mathcal{L}(\theta^*) \in \mathbb{R}^{k}$ denote the class-wise loss vectors for the model parameters at time $t$ and the optimal model parameters. Suppose each class-wise loss $\mathcal{L}_i(\theta) \in \mathbb{R}$ is strongly convex in $\theta$, with strong convexity parameter $\mu_i \in \mathbb{R}_{> 0}$, $\forall i \in \{1,\ldots,k\}$, and each class-wise loss gradient $\smash{\nabla_{\theta} \mathcal{L}_i(\theta) \in \mathbb{R}^m}$ is Lipschitz continuous in $\theta$, having Lipschitz constant $L_i \in \mathbb{R}_{\geq 0}$, $\forall i \in \{1,\ldots,k\}$. Let $\smash{\mu^* = \min_{i \in \{1,\ldots,k \}} \mu_i}$, $\smash{L^* = \max_{i \in \{1,\ldots,k \}} L_i}$. Then, if the model parameters at time $t+1$ are obtained via the gradient of the loss for learn2mix training, where:
\begin{equation}
    \theta^{t+1} = \theta^{t} - \eta \nabla_{\theta} \mathcal{L}(\theta^{t}, \alpha^{t}), \quad \quad \text{with: \quad $\eta \in \mathbb{R}_{>0}$},
\end{equation}
It follows that for learning rate, $\eta \in (0, 2/L^*)$, and for mixing rate, $\gamma \in (0,1)$:
\begin{equation}
    \lim_{t \rightarrow \infty} \theta^t = \theta^*, \quad \quad \text{and:} \quad \lim_{t \rightarrow \infty} \alpha^t = \alpha^* = \frac{\mathcal{L}(\theta^*)}{\mathds{1}_k^T \mathcal{L}(\theta^*)}.
\end{equation}
\end{proposition1}
\begin{proof}
We begin by recalling that $\mathcal{L}_i(\theta)$ is strongly convex in $\theta$ with strong convexity parameter $\mu_i$, $\forall i \in \{1,\ldots,k\}$. Accordingly, $\smash{\forall \alpha \in [0,1]^k}$, with $\smash{\sum_{i=1}^{\raisebox{-0.25ex}{$\scriptstyle k$}} \alpha_i = 1}$, the loss function $\mathcal{L}(\theta,\alpha)$ is strongly convex in $\theta$ with parameter, $\mu' \in \mathbb{R}_{>0}$, which is lower bounded by $\mu^* \in \mathbb{R}_{>0}$, as per Eq.~(\ref{eq:convexity_theta}).
\begin{equation} \label{eq:convexity_theta}
    \mu' \geq \mu^* > 0, \quad \quad \text{where:} \quad \mu^* = \min_{i \in \{1,\ldots,k \}} \mu_i, \quad \quad \text{and:} \quad \mu' = \sum\nolimits_{i=1}^k \alpha_i \mu_i.
\end{equation}
We note that this lower bound on the strong convexity parameter, $\mu' \geq \mu^*$, holds independently of $\alpha$. Now, recall that $\nabla_{\theta} \mathcal{L}_i(\theta)$, is Lipschitz continuous in $\theta$ with Lipschitz constant $L_i$, $\forall i \in \{1,\ldots,k\}$. Accordingly, $\smash{\forall \alpha \in [0,1]^k}$, where $\smash{\sum_{i=1}^{\raisebox{-0.25ex}{$\scriptstyle k$}} \alpha_i = 1}$, the loss gradient $\nabla_{\theta} \mathcal{L}(\theta,\alpha)$ is Lipschitz continuous in $\theta$ with Lipschitz constant, $L' \in \mathbb{R}_{\geq 0}$, which is upper bounded by $L^* \in \mathbb{R}_{\geq 0}$, as per Eq.~(\ref{eq:grad_lipschitz_theta}).
\begin{equation} \label{eq:grad_lipschitz_theta}
    L^* \geq L' \geq 0, \quad \quad \text{where:} \quad L^* = \max_{i \in \{1,\ldots,k \}} L_i, \quad \quad \text{and:} \quad L' = \sum\nolimits_{i=1}^k \alpha_i L_i.
\end{equation}
We affirm that this upper bound on the Lipschitz constant, $L' \leq L^*$, holds independently of $\alpha$. Now, suppose that $\alpha = \alpha^t$, where $\mathcal{L}(\theta,\alpha^t)$ is strongly convex in $\theta$ with parameter $\mu' \geq \mu^*$ and $\nabla_{\theta} \mathcal{L}(\theta,\alpha^t)$ is Lipschitz continuous in $\theta$ with constant $L' \leq L^*$. Let $\rho = \max \{|1 - \eta \mu^*|, |1 - \eta L^*|\}$. By the gradient descent convergence theorem, for learning rate, $\eta \in (0,2/L^*)$, it follows that:
\begin{align}
    \lim_{t \rightarrow \infty} \| \theta^t - \theta^*\| &\leq \lim_{t \rightarrow \infty} \rho^t \| \theta^0 - \theta^*\| = \| \theta^0 - \theta^*\| \lim_{t \rightarrow \infty} \rho^t = 0.
\end{align}
Therefore, $\lim_{t \rightarrow \infty} \theta^t = \theta^*$. Let $\smash{\beta^{t-1} = \mathcal{L}(\theta^{t-1})/\big[\mathds{1}_k^T \mathcal{L}(\theta^{t-1})\big]}$, wherein $\beta^{t-1} \in [0,1]^k$. Unrolling the recurrence relation from Eq.~(\ref{eq:mixing_update}) and expressing it in terms of $\beta^{t-1}$, we obtain:
\begin{align}
    \alpha^t = (1 - \gamma)^t \alpha^0 + \gamma \sum_{l=0}^{t-1}(1 - \gamma)^{t-1-l}\beta^l.
\end{align}
Taking the limit and re-indexing the summation using $n = t-1-l$ and $l = t-1-n$, we obtain:
\begin{align}
    \lim_{t \rightarrow \infty} \alpha^t &= \lim_{t \rightarrow \infty} \Big[ (1 - \gamma)^t \alpha^0 \Big] + \lim_{t \rightarrow \infty} \Bigg[ \gamma \sum_{n=0}^{t-1}(1 - \gamma)^{n}\beta^{t-1-n} \Bigg] \\ \label{eq:reindexed_lims}
    &= \mathbf{0}_k + \gamma \lim_{t \rightarrow \infty} \Bigg[ \sum_{n=0}^{t-1}(1 - \gamma)^{n}\beta^{t-1-n} \Bigg].
\end{align}
We proceed with the steps to invoke the dominated convergence theorem. We note that for fixed $n$:
\begin{align}
    \lim_{t \rightarrow \infty} \Big[ (1 - \gamma)^{n}\beta^{t-1-n} \Big] = (1 - \gamma)^{n} \lim_{t \rightarrow \infty} \Bigg[\frac{\mathcal{L}(\theta^{t-1})}{\mathds{1}_k^T \mathcal{L}(\theta^{t-1})} \Bigg] = (1 - \gamma)^{n} \frac{\mathcal{L}(\theta^*)}{\mathds{1}_k^T \mathcal{L}(\theta^*)}.
\end{align}
Now, consider $g(n) = (1-\gamma)^n$. For this choice of $g(n)$, we have that:
\begin{align}
    \| (1 - \gamma)^{n}\beta^{t-1-n} \| &\leq (1 - \gamma)^{n} \| \beta^{t-1-n} \| \leq g(n), \ \forall t,n \in \mathbb{N} \\
    \sum_{n=0}^\infty g(n) &= \sum_{n=0}^\infty (1 - \gamma)^{n} = \frac{1}{1 - (1 - \gamma)} = \frac{1}{\gamma} < \infty.
\end{align}
We now invoke the dominated convergence theorem. Recalling Eq.~(\ref{eq:reindexed_lims}), we observe that:
\begin{align}
    \lim_{t \rightarrow \infty} \alpha^t &= \gamma \lim_{t \rightarrow \infty} \Bigg[ \sum_{n=0}^{t-1}(1 - \gamma)^{n}\beta^{t-1-n} \Bigg] \\
    &= \gamma \sum_{n=0}^\infty (1 - \gamma)^n \lim_{t \rightarrow \infty} \beta^{t-1-n} = \gamma \sum_{n=0}^\infty (1 - \gamma)^n \frac{\mathcal{L}(\theta^*)}{\mathds{1}_k^T \mathcal{L}(\theta^*)} \\
    &= (\gamma) \bigg(\frac{1}{\gamma}\bigg) \frac{\mathcal{L}(\theta^*)}{\mathds{1}_k^T \mathcal{L}(\theta^*)} = \frac{\mathcal{L}(\theta^*)}{\mathds{1}_k^T \mathcal{L}(\theta^*)} = \alpha^*.
\end{align}
Therefore, $\lim_{t \rightarrow \infty} \alpha^t = \alpha^* = \mathcal{L}(\theta^*)/\big[\mathds{1}_k^T \mathcal{L}(\theta^*)\big]$. Cumulatively, for $\eta \in (0, 2/L^*)$ and $\gamma \in (0,1)$, under learn2mix training, $\lim_{t \rightarrow \infty} \theta^t = \theta^*$, and $\smash{\lim_{t \rightarrow \infty} \alpha^t = \alpha^* = \mathcal{L}(\theta^*)/\big[\mathds{1}_k^T \mathcal{L}(\theta^*)\big]}$.
\end{proof}

\begin{corollary1}
Let $\mathcal{L}(\theta^t) \in \mathbb{R}^k$ denote the class-wise loss vector at time $t$. Suppose each class-wise loss, $\mathcal{L}_i(\theta) \in \mathbb{R}$, is strongly convex in $\theta$, with strong convexity parameter $\mu_i \in \mathbb{R}_{> 0}$, $\forall i \in \{1,\ldots,k\}$, and suppose each class-wise loss gradient $\smash{\nabla_{\theta} \mathcal{L}_i(\theta) \in \mathbb{R}^m}$ is Lipschitz continuous in $\theta$ with Lipschitz constant $L_i \in \mathbb{R}_{\geq 0}$, $\forall i \in \{1,\ldots,k\}$. Let $\smash{\mu^* = \min_{i \in \{1,\ldots,k \}} \mu_i}$, $\smash{L^* = \max_{i \in \{1,\ldots,k \}} L_i}$. Then, the following holds, $\forall \alpha \in [0,1]^k$, with $\smash{\sum_{i=1}^{\raisebox{-0.25ex}{$\scriptstyle k$}} \alpha_i = 1}$:
\begin{align}
    &\frac{\mu^*}{2} \| \theta^t - \theta^* \| \leq \| \nabla_{\theta} \mathcal{L}(\theta^t, \alpha) \| \leq L^* \| \theta^t - \theta^* \|, \\
    &\text{Wherein: } \| \nabla_{\theta} \mathcal{L}(\theta^t, \alpha^t) \| + \| \nabla_{\theta} \mathcal{L}(\theta^t, \tilde{\alpha}) \| \leq 2L^* \| \theta^t - \theta^* \|.
\end{align}
\end{corollary1}
\begin{proof}
We begin by recalling that $\mathcal{L}_i(\theta)$ is strongly convex in $\theta$ with strong convexity parameter $\mu_i$, $\forall i \in \{1,\ldots,k\}$. Accordingly, $\smash{\forall \alpha \in [0,1]^k}$, with $\smash{\sum_{i=1}^{\raisebox{-0.25ex}{$\scriptstyle k$}} \alpha_i = 1}$, the loss function $\mathcal{L}(\theta,\alpha)$ is strongly convex in $\theta$ with parameter, $\mu' \in \mathbb{R}_{>0}$, which is lower bounded by $\mu^* \in \mathbb{R}_{>0}$, as per Eq.~(\ref{eq:convexity_theta_corr}).
\begin{equation} \label{eq:convexity_theta_corr}
    \mu' \geq \mu^* > 0, \quad \quad \text{where:} \quad \mu^* = \min_{i \in \{1,\ldots,k \}} \mu_i, \quad \quad \text{and:} \quad \mu' = \sum\nolimits_{i=1}^k \alpha_i \mu_i.
\end{equation}
Now, recall that $\nabla_{\theta} \mathcal{L}_i(\theta)$, is Lipschitz continuous in $\theta$ with Lipschitz constant $L_i$, $\forall i \in \{1,\ldots,k\}$. Accordingly, $\smash{\forall \alpha \in [0,1]^k}$, where $\smash{\sum_{i=1}^{\raisebox{-0.25ex}{$\scriptstyle k$}} \alpha_i = 1}$, the loss gradient $\nabla_{\theta} \mathcal{L}(\theta,\alpha)$ is Lipschitz continuous in $\theta$ with Lipschitz constant, $L' \in \mathbb{R}_{\geq 0}$, which is upper bounded by $L^* \in \mathbb{R}_{\geq 0}$, as per Eq.~(\ref{eq:grad_lipschitz_theta_corr}).
\begin{equation} \label{eq:grad_lipschitz_theta_corr}
    L^* \geq L' \geq 0, \quad \quad \text{where:} \quad L^* = \max_{i \in \{1,\ldots,k \}} L_i, \quad \quad \text{and:} \quad L' = \sum\nolimits_{i=1}^k \alpha_i L_i.
\end{equation}
Note that $\nabla_\theta \mathcal{L}(\theta^*,\alpha) = \mathbf{0}_m$. Since $\mathcal{L}(\theta,\alpha)$ is strongly convex in $\theta$, the following inequalities hold:
\begin{align} \label{eq:strong_convexity_lower}
    \mathcal{L}(\theta^t,\alpha) - \mathcal{L}(\theta^*,\alpha) &\geq \nabla_\theta \mathcal{L}(\theta^*,\alpha)^T(\theta^t - \theta^*) + \frac{\mu'}{2}\| \theta^t - \theta^* \|^2 = \frac{\mu'}{2}\| \theta^t - \theta^* \|^2, \\ \label{eq:strong_convexity_upper}
    \mathcal{L}(\theta^t,\alpha) - \mathcal{L}(\theta^*,\alpha) &\leq \nabla_\theta \mathcal{L}(\theta^t,\alpha)^T(\theta^t - \theta^*) \leq \| \nabla_\theta \mathcal{L}(\theta^t,\alpha) \| \| \theta^t - \theta^* \|.
\end{align}
Combining Eq.~(\ref{eq:strong_convexity_lower}) and Eq.~(\ref{eq:strong_convexity_upper}), and recalling Eq.~(\ref{eq:convexity_theta_corr}), we obtain the following inequality:
\begin{align} \label{eq:grad_lb}
    \|\nabla_\theta \mathcal{L}(\theta^t,\alpha) \| \geq \frac{\mathcal{L}(\theta^t,\alpha) - \mathcal{L}(\theta^*,\alpha)}{\| \theta^t - \theta^* \|} \geq \frac{\mu^*}{2} \| \theta^t - \theta^* \|.
\end{align}
Furthermore, since $\nabla_{\theta} \mathcal{L}(\theta,\alpha)$ is Lipschitz continuous in $\theta$ and recalling Eq.~(\ref{eq:grad_lipschitz_theta_corr}), it follows that:
\begin{align} \label{eq:grad_ub}
    \|\nabla_\theta \mathcal{L}(\theta^t,\alpha) - \nabla_\theta \mathcal{L}(\theta^*,\alpha) \| \leq L' \| \theta^t - \theta^* \| \implies  \|\nabla_\theta \mathcal{L}(\theta^t,\alpha) \| \leq L^* \| \theta^t - \theta^* \|.
\end{align}
Altogether, combining Eq.~(\ref{eq:grad_lb}) and Eq.~(\ref{eq:grad_ub}), we arrive at the final inequality:
\begin{align}
    \frac{\mu^*}{2} \| \theta^t - \theta^* \| \leq \| \nabla_{\theta} \mathcal{L}(\theta^t, \alpha) \| \leq L^* \| \theta^t - \theta^* \|.
\end{align}
Furthermore, since Eq.~(\ref{eq:grad_ub}) holds $\forall \alpha \in [0,1]^k$ where $\smash{\sum_{i=1}^{\raisebox{-0.25ex}{$\scriptstyle k$}} \alpha_i = 1}$, it follows that:
\begin{align}
    \|\nabla_\theta \mathcal{L}(\theta^t,\alpha^t) \| + \|\nabla_\theta \mathcal{L}(\theta^t,\tilde{\alpha}) \| \leq 2L^* \| \theta^t - \theta^* \|.
\end{align}
\end{proof}

\begin{proposition2}
Let $\mathcal{L}(\theta^t), \mathcal{L}(\theta^*) \in \mathbb{R}^{k}$ denote the respective class-wise loss vectors for the model parameters at time $t$ and for the optimal model parameters. Suppose each class-wise loss, $\mathcal{L}_i(\theta) \in \mathbb{R}$ is strongly convex in $\theta$ with strong convexity parameter $\mu_i \in \mathbb{R}_{> 0}$, $\forall i \in \{1,\ldots,k\}$, and each class-wise loss gradient $\smash{\nabla_{\theta} \mathcal{L}_i(\theta) \in \mathbb{R}^m}$ is Lipschitz continuous in $\theta$, having Lipschitz constant $L_i \in \mathbb{R}_{\geq 0}$, $\forall i \in \{1,\ldots,k\}$. Moreover, suppose that the loss gradient $\smash{\nabla_{\theta} \mathcal{L}(\theta, \alpha) \in \mathbb{R}^m}$ is Lipschitz continuous in $\alpha$, having Lipschitz constant $L_\alpha \in \mathbb{R}_{\geq 0}$, and let $\smash{\mu^* = \min_{i \in \{1,\ldots,k \}} \mu_i}$, $\smash{L^* = \max_{i \in \{1,\ldots,k \}} L_i}$. Then, if and only if the following holds:
\begin{align}
    \Big[\Big(\frac{\mu^*}{2} - L^*\Big)\| \theta^t - \theta^* \|^2 + \tilde{\alpha}^T(\mathcal{L}(\theta^t) - \mathcal{L}(\theta^*)) \Big] \Big[\| \theta^t - \theta^* \| - (\mathcal{L}(\theta^t) - \mathcal{L}(\theta^*))\Big] > 0,
\end{align}
It follows that for every learning rate, $\eta > 0$, and for every mixing rate, $\gamma \in (0,\beta]$:
\begin{align}
    &\left\| \left(\theta^t - \eta \nabla_{\theta}\mathcal{L}(\theta^t, \alpha^t)\right) - \theta^{*} \right\| \leq \left\| \left(\theta^t - \eta \nabla_{\theta}\mathcal{L}(\theta^t, \tilde{\alpha})\right)  - \theta^{*} \right\|, \\
    &\text{Where: } \beta = \frac{ \big(\frac{\mu^*}{2} - L^*\big)\| \theta^t - \theta^* \|^2 + \tilde{\alpha}^T(\mathcal{L}(\theta^t) - \mathcal{L}(\theta^*))}{ \eta L_\alpha L^* \Big\| \frac{\mathcal{L}(\theta^{t-1})}{\mathds{1}_k^T \mathcal{L}(\theta^{t-1})} - \tilde{\alpha} \Big\| \Big[ \| \theta^t - \theta^* \| - (\mathcal{L}(\theta^t) - \mathcal{L}(\theta^*)) \Big] }
\end{align}

\end{proposition2}
\begin{proof}
We note that for all subsequent derivations, $\mathcal{F}(\theta^t,\theta^*,\eta,\alpha^t) =  \left\| \left(\theta^t - \eta \nabla_{\theta}\mathcal{L}(\theta^t, \alpha^t)\right) - \theta^{*} \right\|$, and $\mathcal{G}(\theta^t,\theta^*,\eta,\tilde{\alpha}) = \left\| \left(\theta^t - \eta \nabla_{\theta}\mathcal{L}(\theta^t, \tilde{\alpha})\right)  - \theta^{*} \right\|$, where $\alpha^{t-1} = \tilde{\alpha}$. We begin by observing that:
\begin{align}
    \big[\mathcal{F}(\theta^t,\theta^*,\eta,\alpha^t)\big]^2 &= \| \theta^t - \theta^* \| ^2 - 2\eta(\theta^t - \theta^*)^T \nabla_{\theta}\mathcal{L}(\theta^t, \alpha^t) + \eta^2 \| \nabla_{\theta}\mathcal{L}(\theta^t, \alpha^t) \|^2, \\
    \big[\mathcal{F}(\theta^t,\theta^*,\eta,\tilde{\alpha})\big]^2 &= \| \theta^t - \theta^* \| ^2 - 2\eta(\theta^t - \theta^*)^T \nabla_{\theta}\mathcal{L}(\theta^t, \tilde{\alpha}) + \eta^2 \| \nabla_{\theta}\mathcal{L}(\theta^t, \tilde{\alpha}) \|^2.
\end{align}
Accordingly, the difference between $\big[\mathcal{F}(\theta^t,\theta^*,\eta,\alpha^t)\big]^2$ and $\big[\mathcal{G}(\theta^t,\theta^*,\eta,\tilde{\alpha})\big]^2$ is given by:
\begin{align}
    \begin{split}
    \big[\mathcal{F}(\theta^t,\theta^*,\eta,\alpha^t)\big]^2 - \big[\mathcal{G}(\theta^t,\theta^*,\eta,\tilde{\alpha})\big]^2 &= - 2\eta \big[(\theta^t - \theta^*)^T(\nabla_{\theta}\mathcal{L}(\theta^t, \alpha^t) - \nabla_{\theta}\mathcal{L}(\theta^t, \tilde{\alpha})) \big] \\
    &\qquad \ + \eta^2 \big[ \| \nabla_{\theta} \mathcal{L}(\theta^t, \alpha^t) \|^2 - \| \nabla_{\theta} \mathcal{L}(\theta^t, \tilde{\alpha}) \|^2 \big].
    \end{split}
\end{align}
Consequently, suppose that $\mathcal{H}(\theta^t,\theta^*,\eta,\tilde{\alpha},\alpha^t) = 2\eta \big[(\theta^t - \theta^*)^T(\nabla_{\theta}\mathcal{L}(\theta^t, \alpha^t) - \nabla_{\theta}\mathcal{L}(\theta^t, \tilde{\alpha})) \big]$, and let $\mathcal{J}(\theta^t,\eta,\tilde{\alpha},\alpha^t) = \eta^2 \big[ \| \nabla_{\theta} \mathcal{L}(\theta^t, \alpha^t) \|^2 - \| \nabla_{\theta} \mathcal{L}(\theta^t, \tilde{\alpha}) \|^2 \big]$. Suppose the loss gradient, $\smash{\nabla_{\theta} \mathcal{L}(\theta, \alpha)}$, is Lipschitz continuous in $\alpha$ with Lipschitz constant, $L_\alpha$. We now upper bound $\mathcal{J}(\theta^t,\eta,\tilde{\alpha},\alpha^t)$: 
\begin{align}
    \notag \mathcal{J}(\theta^t,\eta,\alpha,\alpha^t) &= \eta^2 \big[\nabla_{\theta} \mathcal{L}(\theta^t,\alpha^t) - \nabla_{\theta} \mathcal{L}(\theta^t,\tilde{\alpha})\big]^T \big[\nabla_{\theta} \mathcal{L}(\theta^t,\alpha^t) + \nabla_{\theta} \mathcal{L}(\theta^t,\tilde{\alpha})\big] \\
    &\leq \| \nabla_{\theta} \mathcal{L}(\theta^t,\alpha^t) - \nabla_{\theta} \mathcal{L}(\theta^t,\tilde{\alpha}) \| \| \nabla_{\theta} \mathcal{L}(\theta^t,\alpha^t) + \nabla_{\theta} \mathcal{L}(\theta^t,\tilde{\alpha}) \| \\
    &\leq 2 \eta^2 L_\alpha \| \alpha^t - \tilde{\alpha} \| \Big[ \| \nabla_{\theta} \mathcal{L}(\theta^t,\alpha^t) \| + \| \nabla_{\theta} \mathcal{L}(\theta^t,\tilde{\alpha}) \| \Big] \\
    &\leq 2 \eta^2 L_\alpha L^* \| \alpha^t - \tilde{\alpha} \| \| \theta^t - \theta^* \| \\
    &= 2 \eta^2 L_\alpha L^* \bigg\| \tilde{\alpha} + \gamma \bigg(\frac{\mathcal{L}(\theta^{t-1})}{\mathds{1}_k^T \mathcal{L}(\theta^{t-1})} - \tilde{\alpha} \bigg) - \tilde{\alpha} \bigg\| \| \theta^t - \theta^* \| \\ \label{eq:convergence_lb}
    &= 2 \eta^2 L_\alpha L^* \gamma \bigg\| \frac{\mathcal{L}(\theta^{t-1})}{\mathds{1}_k^T \mathcal{L}(\theta^{t-1})} - \tilde{\alpha} \bigg\| \| \theta^t - \theta^* \|.
\end{align}
We note that this upper bound follows from the Cauchy-Schwarz inequality and Corollary \ref{corr:loss_and_grad}. We now proceed by lower bounding $\mathcal{H}(\theta^t,\theta^*,\eta,\tilde{\alpha},\alpha^t)$:
\begin{align}
    \mathcal{H}(\theta^t,\theta^*,\eta,\tilde{\alpha},\alpha^t) &= 2\eta \Big[ (\theta^t - \theta^*)^T \nabla_{\theta} \mathcal{L}(\theta^t, \alpha^t) - (\theta^t - \theta^*)^T \nabla_{\theta} \mathcal{L}(\theta^t, \tilde{\alpha}) \Big] \\
    &\geq 2\eta \Big[ (\theta^t - \theta^*)^T \nabla_{\theta} \mathcal{L}(\theta^t, \alpha^t) - \| \theta^t - \theta^* \| \| \nabla_{\theta} \mathcal{L}(\theta^t, \tilde{\alpha}) \| \Big] \\
    &\geq 2\eta \Big[ (\theta^t - \theta^*)^T \nabla_{\theta} \mathcal{L}(\theta^t, \alpha^t) - L^* \| \theta^t - \theta^* \|^2 \Big] \\
    &= 2 \eta \bigg[ \frac{\mu^*}{2} \| \theta^t - \theta^* \|^2 + \mathcal{L}(\theta^t, \alpha^t) - \mathcal{L}(\theta^*, \alpha^t) - L^* \| \theta^t - \theta^* \|^2 \bigg] \\ \label{eq:convergence_ub}
    \notag &= 2 \eta \bigg[ \Big( \frac{\mu^*}{2} - L^* \Big) \| \theta^t - \theta^* \|^2 + \tilde{\alpha}^T(\mathcal{L}(\theta^t) - \mathcal{L}(\theta^*)) \\
    &\qquad \quad + \gamma \bigg( \frac{\mathcal{L}(\theta^{t-1})}{\mathds{1}^T \mathcal{L}(\theta^{t-1})} - \tilde{\alpha} \bigg)^T (\mathcal{L}(\theta^t) - \mathcal{L}(\theta^*)) \bigg].
\end{align}
As in above, we note that this lower bound also follows from the Cauchy-Schwarz inequality and Corollary \ref{corr:loss_and_grad}, and further invokes the strong convexity of $\mathcal{L}(\theta,\alpha)$ in $\theta$. Combining Eq.~(\ref{eq:convergence_lb}) and Eq.~(\ref{eq:convergence_ub}), we obtain the following upper bound on $\smash{[\mathcal{F}(\theta^t,\theta^*,\eta,\alpha^t)]^2 - [\mathcal{G}(\theta^t,\theta^*,\eta,\tilde{\alpha})]^2 }$:
\begin{align} \label{eq:combined_bounds}
    &\big[\mathcal{F}(\theta^t,\theta^*,\eta,\alpha^t)\big]^2 - \big[\mathcal{G}(\theta^t,\theta^*,\eta,\tilde{\alpha})\big]^2 \leq \mathcal{K}(\theta^t,\theta^*,\eta,\gamma,\tilde{\alpha},\alpha^t), \\
    \notag &\text{Where: } \mathcal{K}(\theta^t,\theta^*,\eta,\gamma,\tilde{\alpha},\alpha^t) = - 2 \eta \bigg[ \Big( \frac{\mu^*}{2} - L^* \Big) \| \theta^t - \theta^* \|^2 + \tilde{\alpha}^T(\mathcal{L}(\theta^t) - \mathcal{L}(\theta^*)) \\
    \notag &\qquad \qquad \qquad \qquad \qquad \qquad \qquad \quad + \gamma \bigg( \frac{\mathcal{L}(\theta^{t-1})}{\mathds{1}^T \mathcal{L}(\theta^{t-1})} - \tilde{\alpha} \bigg)^T (\mathcal{L}(\theta^t) - \mathcal{L}(\theta^*)) \bigg] \\
    &\qquad \qquad \qquad \qquad \qquad \qquad \qquad \quad + 2 \eta^2 L_\alpha L^* \gamma \bigg\| \frac{\mathcal{L}(\theta^{t-1})}{\mathds{1}_k^T \mathcal{L}(\theta^{t-1})} - \tilde{\alpha} \bigg\| \| \theta^t - \theta^* \|.
\end{align}
Now, consider the following chain of inequalities deriving from Eq.~(\ref{eq:combined_bounds}):
\begin{align}
\begin{split}
    \mathcal{K}(\theta^t,\theta^*,\eta,\gamma,\tilde{\alpha},\alpha^t) \leq 0 &\implies \big[\mathcal{F}(\theta^t,\theta^*,\eta,\alpha^t)\big]^2 - \big[\mathcal{G}(\theta^t,\theta^*,\eta,\tilde{\alpha})\big]^2 \leq 0 \\
    &\implies \big[\mathcal{F}(\theta^t,\theta^*,\eta,\alpha^t)\big] \leq \big[\mathcal{G}(\theta^t,\theta^*,\eta,\tilde{\alpha})\big].
\end{split}
\end{align}
Accordingly, we aim to find a condition on the mixing rate, $\gamma$, under which the chain of inequalities is satisfied. We proceed by letting $\mathcal{K}(\theta^t,\theta^*,\eta,\gamma,\tilde{\alpha},\alpha^t) \leq 0$, and rearrange the terms:
\begin{align}
    \Big(\frac{\mu^*}{2} - L^*\Big)\| \theta^t - \theta^* \|^2 + \tilde{\alpha}^T(\mathcal{L}(\theta^t) - \mathcal{L}(\theta^*)) &\geq \gamma \bigg[ \eta L_\alpha L^* \bigg\| \frac{\mathcal{L}(\theta^{t-1})}{\mathds{1}_k^T \mathcal{L}(\theta^{t-1})} - \tilde{\alpha} \bigg\| \| \theta^t - \theta^* \| \\
    \notag &\qquad \quad - \bigg( \frac{\mathcal{L}(\theta^{t-1})}{\mathds{1}^T \mathcal{L}(\theta^{t-1})} - \tilde{\alpha} \bigg)^T (\mathcal{L}(\theta^t) - \mathcal{L}(\theta^*)) \bigg].
\end{align}
We note that this chain of inequalities is satisfied if, for every $\eta > 0$:
\begin{align}
    &0 < \gamma \leq \frac{ \big(\frac{\mu^*}{2} - L^*\big)\| \theta^t - \theta^* \|^2 + \tilde{\alpha}^T(\mathcal{L}(\theta^t) - \mathcal{L}(\theta^*))}{ \eta L_\alpha L^* \Big\| \frac{\mathcal{L}(\theta^{t-1})}{\mathds{1}_k^T \mathcal{L}(\theta^{t-1})} - \tilde{\alpha} \Big\| \| \theta^t - \theta^* \| - \Big( \frac{\mathcal{L}(\theta^{t-1})}{\mathds{1}^T \mathcal{L}(\theta^{t-1})} - \tilde{\alpha} \Big)^T (\mathcal{L}(\theta^t) - \mathcal{L}(\theta^*)) } \leq \beta, \\ \label{eq:num_den_positive}
    &\text{Where: } \beta = \frac{ \big(\frac{\mu^*}{2} - L^*\big)\| \theta^t - \theta^* \|^2 + \tilde{\alpha}^T(\mathcal{L}(\theta^t) - \mathcal{L}(\theta^*))}{ \eta L_\alpha L^* \Big\| \frac{\mathcal{L}(\theta^{t-1})}{\mathds{1}_k^T \mathcal{L}(\theta^{t-1})} - \tilde{\alpha} \Big\| \Big[ \| \theta^t - \theta^* \| - (\mathcal{L}(\theta^t) - \mathcal{L}(\theta^*)) \Big] }.
\end{align}
However, $\gamma > 0$ iff the numerator and denominator from Eq.~(\ref{eq:num_den_positive}) have the same sign, ensuring that $\beta > 0$. Accordingly, if and only if the condition provided in Eq.~(\ref{eq:sign_condition}) is satisfied:
\begin{align} \label{eq:sign_condition}
    \Big[\Big(\frac{\mu^*}{2} - L^*\Big)\| \theta^t - \theta^* \|^2 + \tilde{\alpha}^T(\mathcal{L}(\theta^t) - \mathcal{L}(\theta^*)) \Big] \Big[\| \theta^t - \theta^* \| - (\mathcal{L}(\theta^t) - \mathcal{L}(\theta^*))\Big] > 0,
\end{align}
It follows that for every learning rate $\eta > 0$, and for every mixing rate $\gamma \in (0,\beta]$ satisfying Eq.~(\ref{eq:num_den_positive}):
\begin{align}
\left\| \left(\theta^t - \eta \nabla_{\theta}\mathcal{L}(\theta^t, \alpha^t)\right) - \theta^{*} \right\| \leq \left\| \left(\theta^t - \eta \nabla_{\theta}\mathcal{L}(\theta^t, \tilde{\alpha})\right)  - \theta^{*} \right\|.
\end{align}
\end{proof}

\section{Additional Empirical Results} \label{sec:additional_results}
\begin{figure*}[h!]
    \centering
    \begin{subfigure}{0.48\textwidth}
        \includegraphics[width=\textwidth]{test_cifar10_time.png}
        \caption{CIFAR-10: Large LeNet}
    \end{subfigure}
    \begin{subfigure}{0.48\textwidth}
        \includegraphics[width=\textwidth]{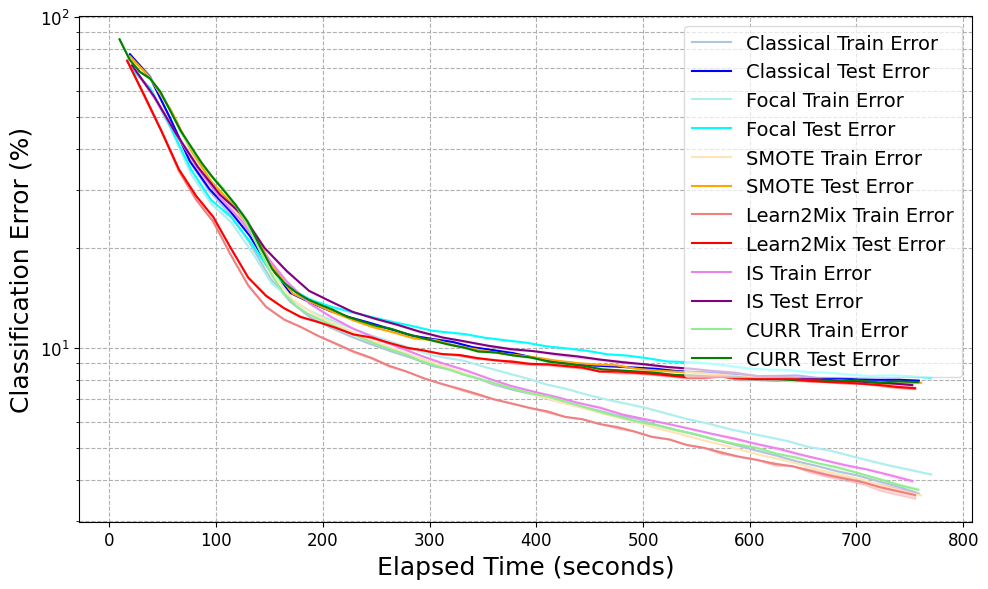}
        \caption{CIFAR-10: MobileNet-V3 Small}
    \end{subfigure}
\\
    \begin{subfigure}{0.48\textwidth}
        \includegraphics[width=\textwidth]{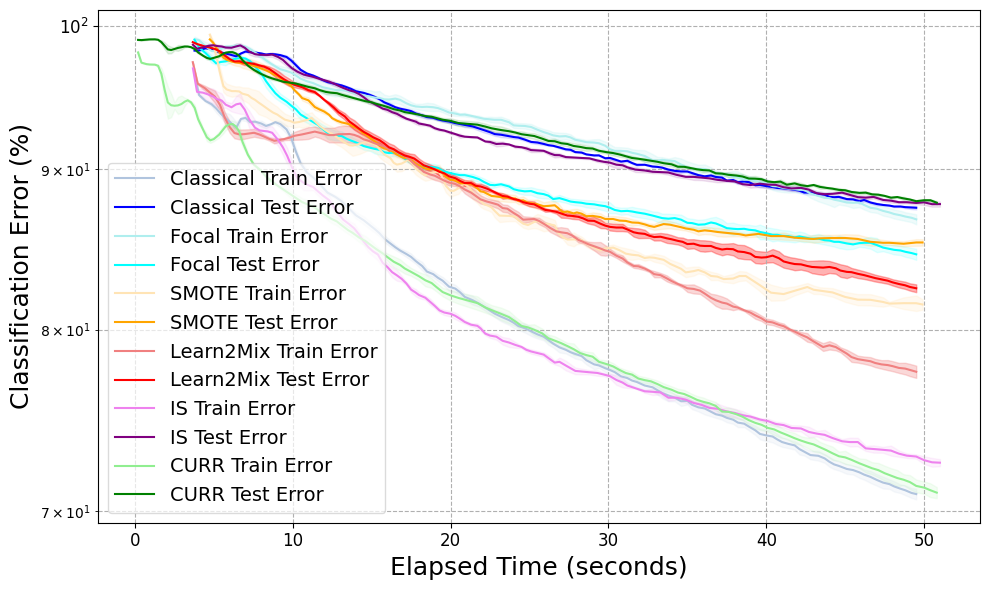}
        \caption{CIFAR-100: Large LeNet}
    \end{subfigure}
    \begin{subfigure}{0.48\textwidth}
        \includegraphics[width=\textwidth]{test_cifar100_time.png}
        \caption{CIFAR-100: MobileNet-V3 Small}
    \end{subfigure}
    \caption{Comparing model classification errors for learn2mix, classical, FCL, SMOTE, IS, and CURR training. The x-axis is the elapsed [training] time, while the y-axis is the classification error.} \label{fig:ablation_architecture}
\end{figure*}

For further performance verification of learn2mix, we present several ablation studies quantifying the effects of different architectures, optimizers, batch sizes, learning rates, and mixing rates for the considered classification tasks from the main text. We further present the worst-class classification accuracy on Imagenette to further gauge the efficacy of learn2mix within imbalanced classification settings, and illustrate how the mixing parameters converge to a stable distribution on Mean Estimation. We first consider CIFAR-10 and CIFAR-100 (per Section \ref{sec:dataset_descriptions}), and evaluate whether the gains afforded by learn2mix persist across architectures. For CIFAR-10, we recall the Large LeNet architecture, trained using the Adam optimizer and Cross Entropy Loss with learning rate $\eta = 7\text{e-}5$ for $\underline{E} = 200$ s, and the MobileNet-V3 Small architecture, trained using the Adam optimizer and Cross Entropy Loss with learning rate $\eta = 1\text{e-}4$ for $\underline{E} = 750$ s. For CIFAR-100, we consider again the MobileNet-V3 Small architecture, trained using the Adam optimizer and Cross Entropy Loss with learning rate $\eta = 1\text{e-}4$ for $\underline{E} = 200$ s, and the Large LeNet architecture, trained using the Adam optimizer and Cross Entropy Loss with learning rate $\eta = 1\text{e-}4$ for $\underline{E} = 50$ s. The results are depicted in Figure \ref{fig:ablation_architecture}. We observe that for both Large LeNet and MobileNet-V3 Small, the learn2mix-trained models converge faster than the classical, FCL, SMOTE, IS, and CURR trained models.

\begin{figure*}[h!]
    \centering
    \begin{subfigure}{0.32\textwidth}
        \includegraphics[width=\textwidth]{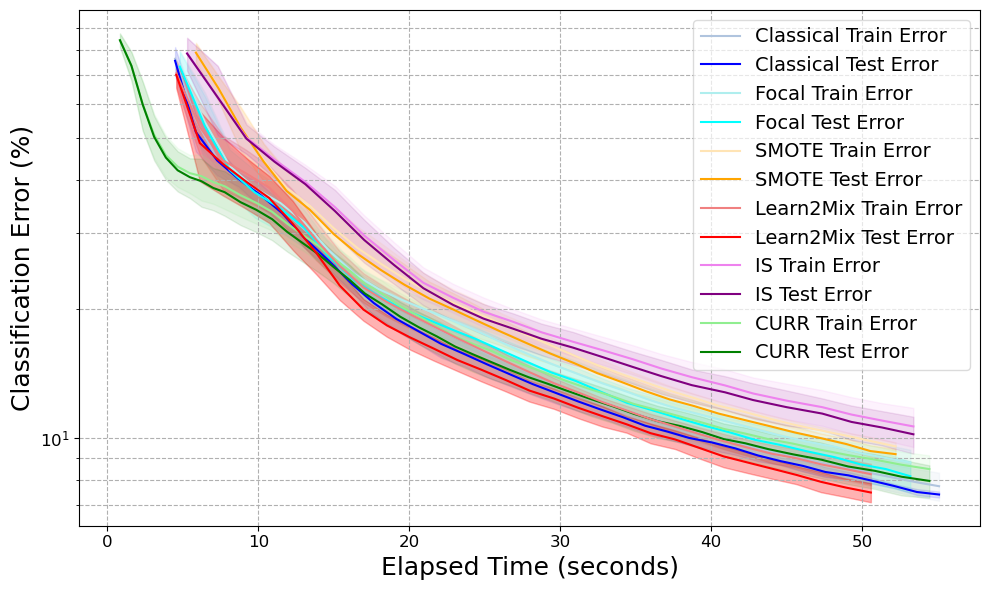}
        \caption{MNIST: $M = 150$}
    \end{subfigure}
    \begin{subfigure}{0.32\textwidth}
        \includegraphics[width=\textwidth]{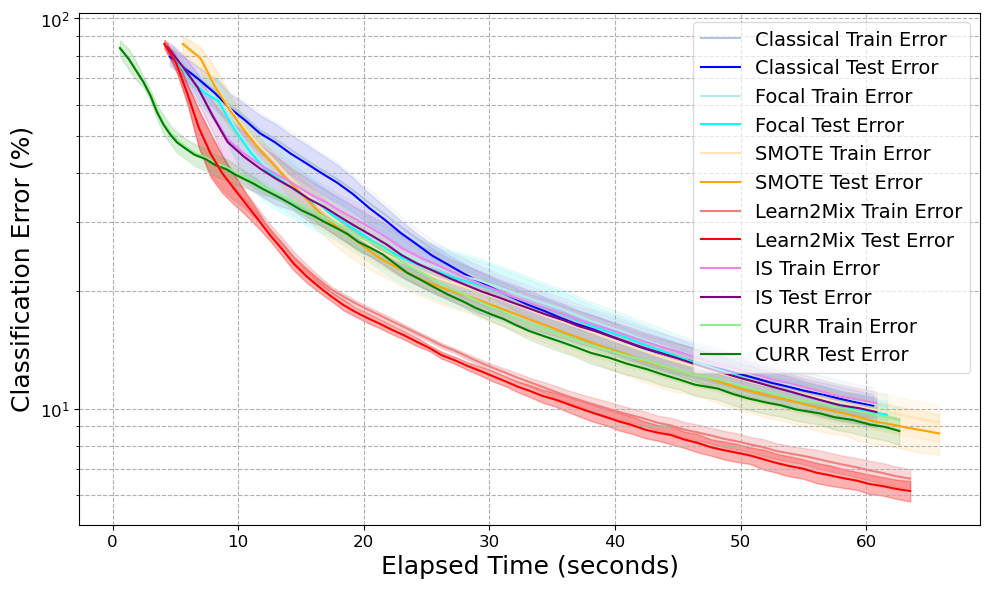}
        \caption{MNIST: $M = 500$}
    \end{subfigure}
    \begin{subfigure}{0.32\textwidth}
        \includegraphics[width=\textwidth]{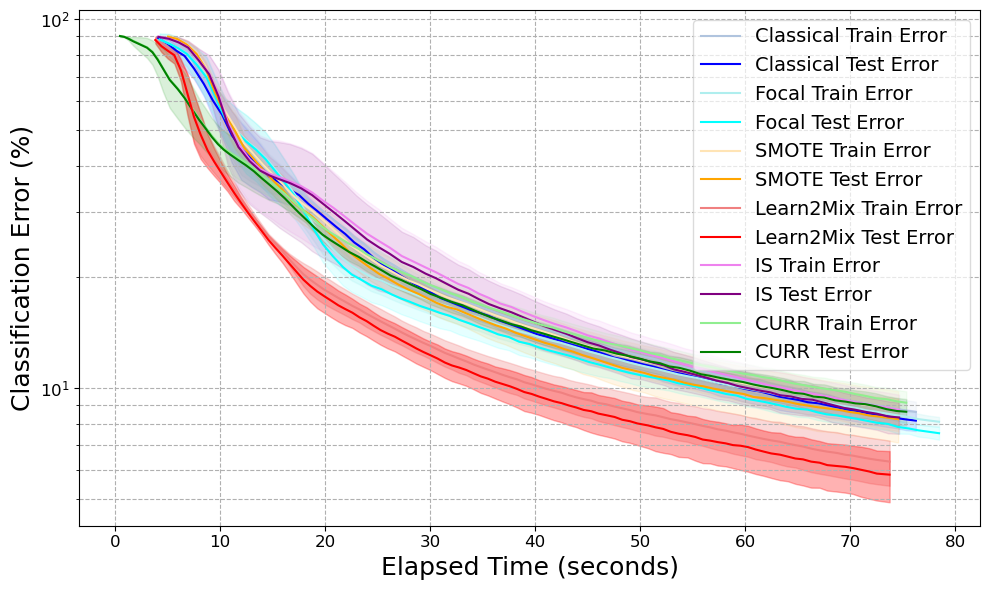}
        \caption{MNIST: $M = 1000$}
    \end{subfigure}
    \caption{Comparing model classification errors for learn2mix, classical, FCL, SMOTE, IS, and CURR training. The x-axis is the elapsed [training] time, while the y-axis is the classification error.} \label{fig:ablation_opt_batch_size}
\end{figure*}

Next, we evaluate the robustness of learn2Mix to different optimizers and batch sizes. As we used the Adam optimizer in the main text, we now consider the RMSProp optimizer \citep{graves2013generating} with batch size $M \in \{250, 500, 1000 \}$. We train LeNet-5 on MNIST using Cross Entropy Loss with learning rate $\eta = 1\text{e-}5$ for $\underline{E} = 45$ s, $\underline{E} = 60$ s, and $\underline{E} = 70$ s. As depicted in Figure \ref{fig:ablation_opt_batch_size}, we see that learn2mix converges faster than classical, FCL, SMOTE, IS, and CURR training.

\begin{figure*}[h!]
    \centering
    \begin{subfigure}{0.32\textwidth}
        \includegraphics[width=\textwidth]{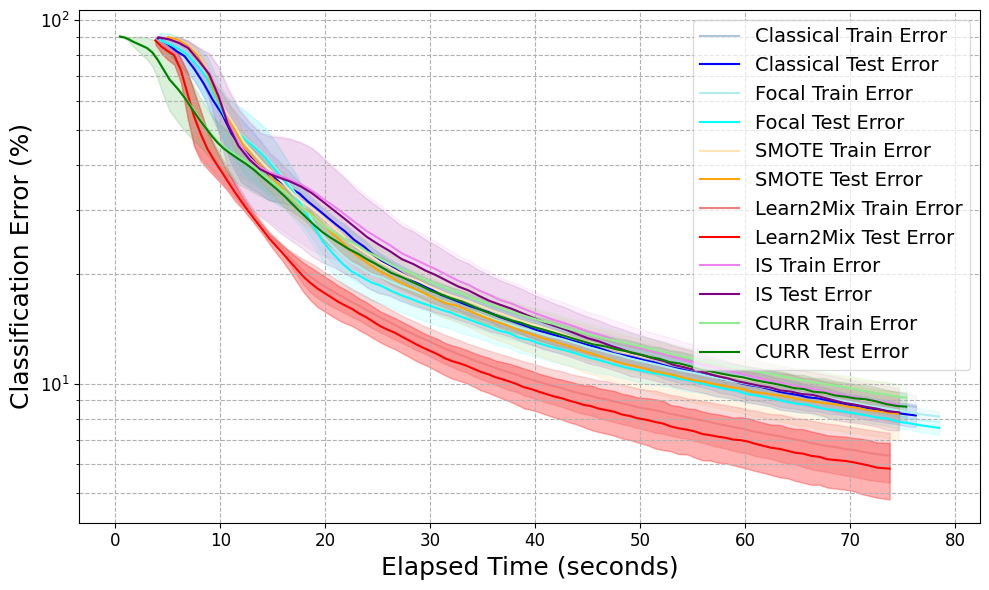}
        \caption{MNIST: $\eta = 1\text{e-}5$}
    \end{subfigure}
    \begin{subfigure}{0.32\textwidth}
        \includegraphics[width=\textwidth]{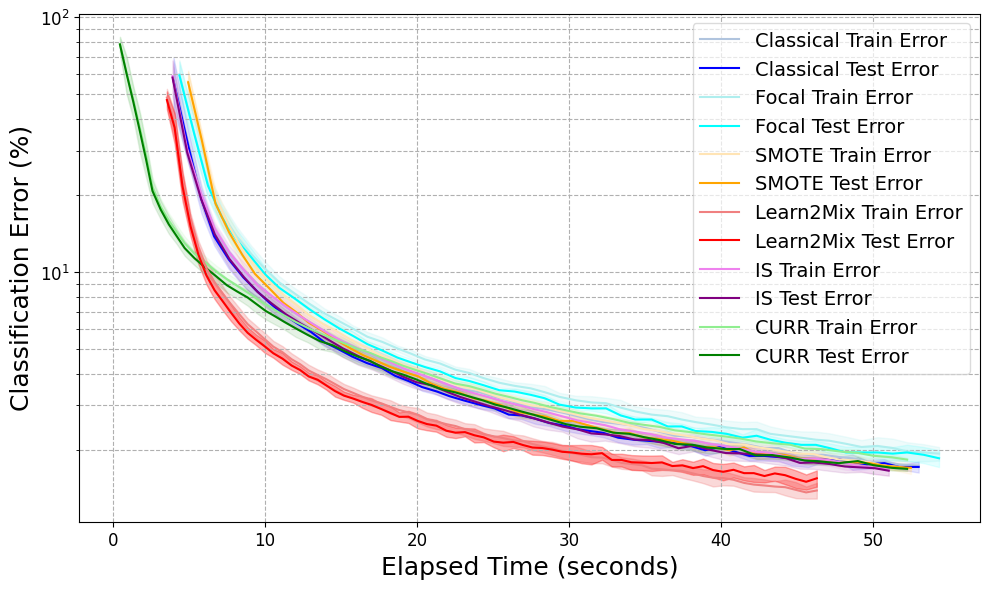}
        \caption{MNIST: $\eta = 1\text{e-}4$}
    \end{subfigure}
    \begin{subfigure}{0.32\textwidth}
        \includegraphics[width=\textwidth]{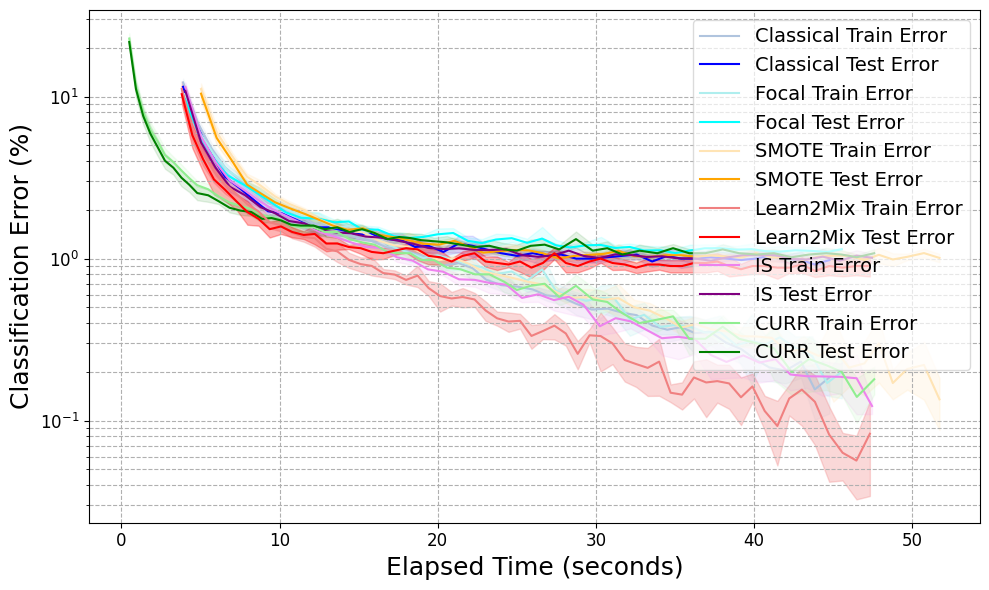}
        \caption{MNIST: $\eta = 1\text{e-}3$}
    \end{subfigure}
    \caption{Comparing model classification errors for learn2mix, classical, FCL, SMOTE, IS, and CURR training. The x-axis is the elapsed [training] time, while the y-axis is the classification error.} \label{fig:ablation_learning_rate}
\end{figure*}

We further verify the robustness of learn2Mix to different learning rates. We train LeNet-5 on MNIST using Cross Entropy Loss with learning rate $\eta \in \{1\text{e-}5, 1\text{e-}4, 1\text{e-}3\}$ for $\underline{E} = 75$ s, $\underline{E} = 50$ s, and $\underline{E} = 45$ s. Per Figure \ref{fig:ablation_learning_rate}, we observe that the faster convergence afforded by learn2mix is apparent for $\eta \in \{ 1\text{e-}5, 1\text{e-}4 \}$. For $\eta = 1\text{e-}3$, we note that after convergence, the learn2mix train error continues to decreases at a faster rate than the the classical, FCL, SMOTE, IS, and CURR train errors.

\begin{figure*}[h!]
    \centering
    \begin{subfigure}{0.48\textwidth}
        \includegraphics[width=\textwidth]{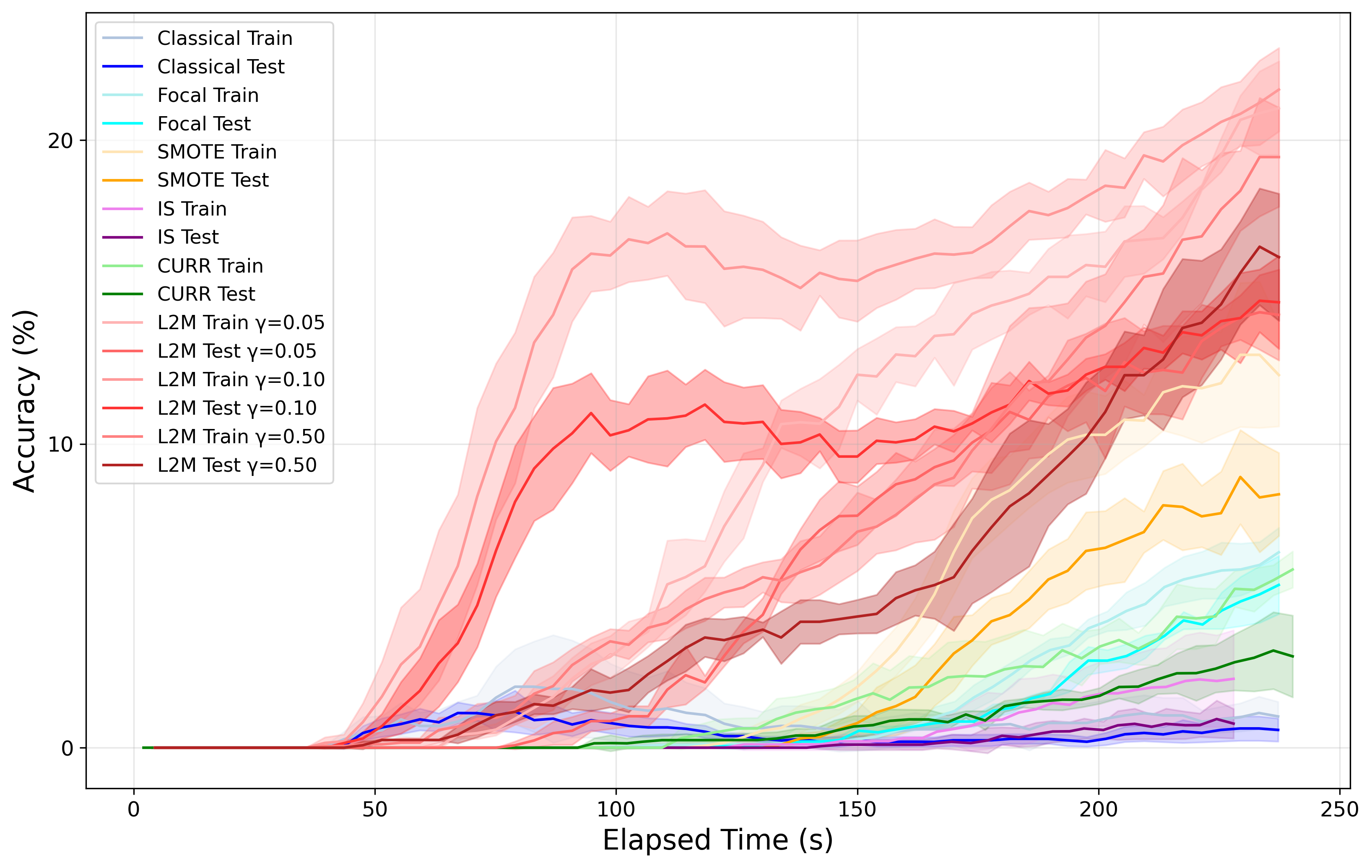}
        \caption{Imagenette}
    \end{subfigure}
    \begin{subfigure}{0.48\textwidth}
        \includegraphics[width=\textwidth]{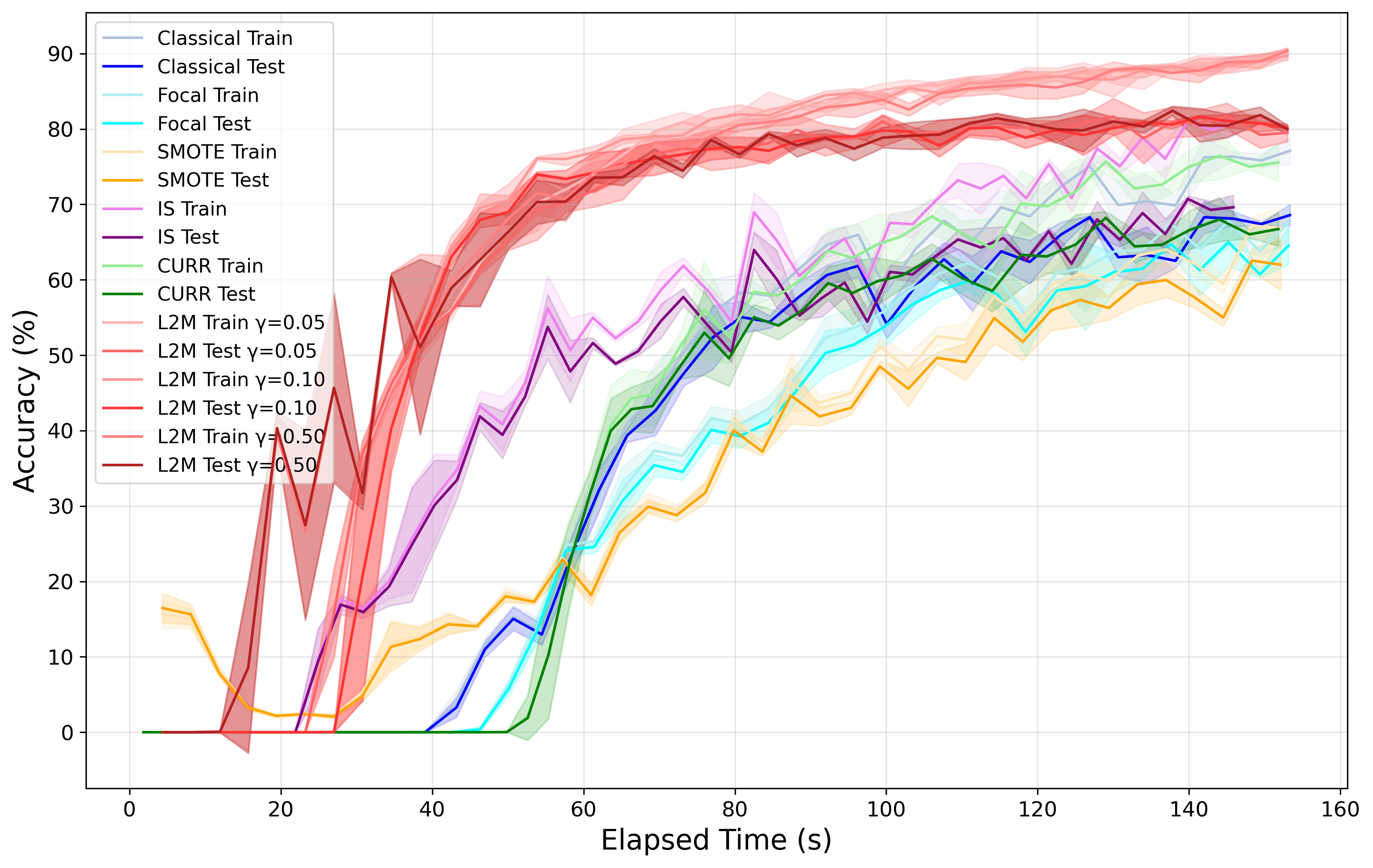}
        \caption{IMDB}
    \end{subfigure}
    \caption{Comparing worst-class model classification accuracies using learn2mix, classical, FCL, SMOTE, IS, and CURR training on Imagenette and IMDB. The x-axis is the elapsed [training] time, while the y-axis is the classification accuracy of the worst-class.}
    \label{fig:ablation_worst-class_acc}
\end{figure*}

We now illustrate the worst-class classification accuracy on Imagenette and IMDB as an additional metric to gauge the efficacy of learn2mix for imbalanced classification settings. We train ResNet-18 on Imagenette via Cross Entropy Loss with learning rate $\eta = 1\text{e-}5$ for $\underline{E} = 240$ s, and a transformer on IMDB using Cross-Entropy Loss with learning rate $\eta = 1\text{e-}4$ for $\underline{E} = 150$ s, and record the test classification accuracy of the worst class after each training epoch, $t$. To demonstrate relative insensitivity to the choice of $\gamma$, we ablate the mixing rate for $\gamma \in [0.01,0.1]$. The result is depicted in Figure \ref{fig:ablation_worst-class_acc}. We see that learn2mix offers a considerable improvement in the worst-class classification accuracy metric versus classical, FCL, SMOTE, IS, and CURR training, which matches intuition; the theoretical foundation of learn2Mix is to increase the proportion of harder classes during training, which directly translates to stronger results for the most challenging classes.

\begin{figure*}[h!]
    \centering
    \includegraphics[width=0.5\linewidth]{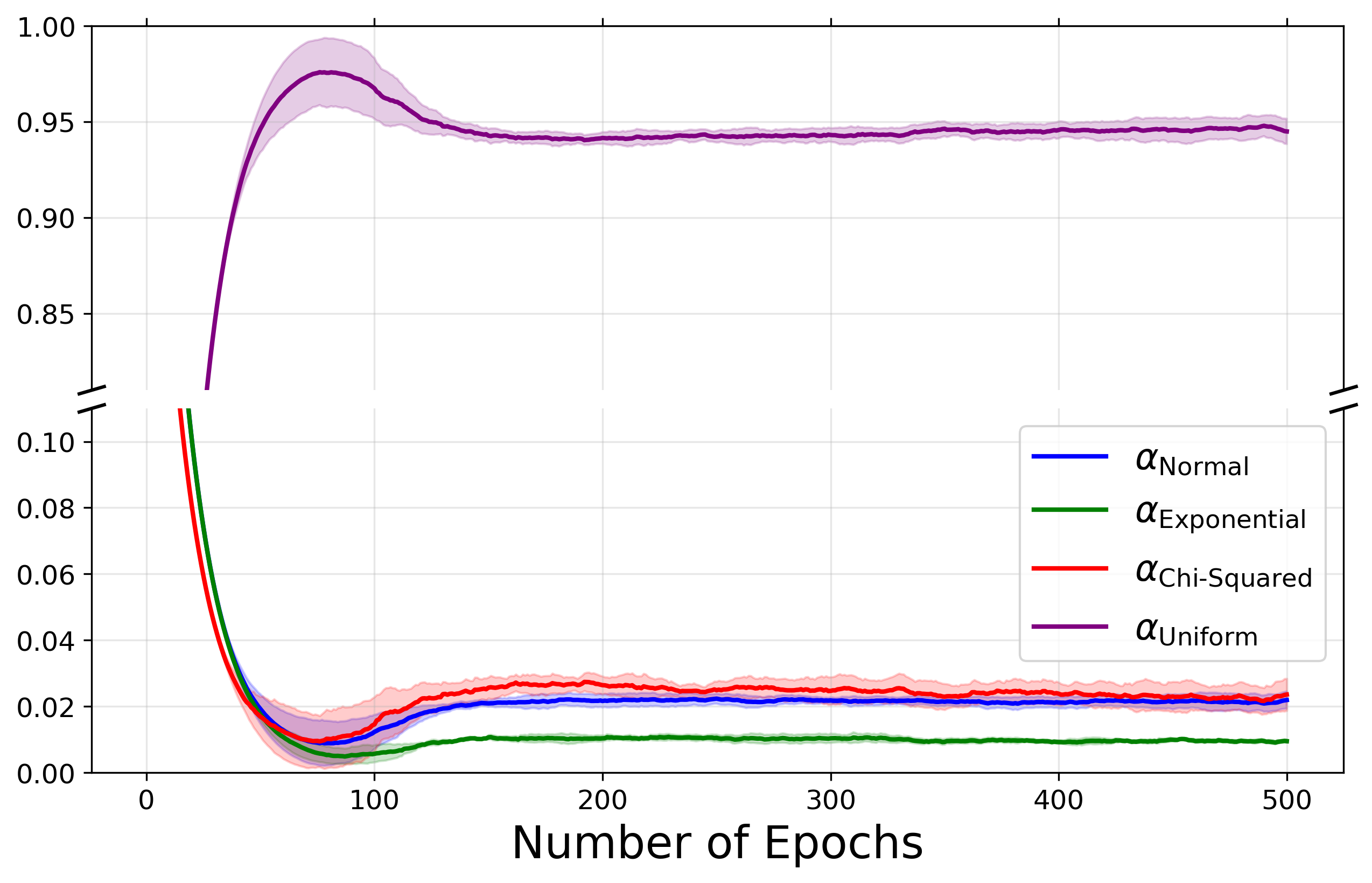}
    \caption{Evolution of learn2mix mixing parameters across training epochs on Mean Estimation.}
    \label{fig:mixing_params_mean_estimation}
\end{figure*}

To illustrate how the mixing parameters converge to a stable distribution during training (as detailed in Section \ref{sec:theoretical_results}), we train a fully connected network on Mean Estimation (where the Normal, Exponential, and Chi-squared cases have similar variance but the Uniform case is substantially more variable) using Cross-Entropy Loss with learning rate $\eta = 5\text{e-}5$ for $E = 500$ epochs. As depicted in Figure \ref{fig:mixing_params_mean_estimation}, learn2mix prioritizes the hardest class without overstating differences among the easier ones.

\section{Dataset Descriptions} \label{sec:dataset_descriptions}
\subsection{MNIST Dataset}
The \textbf{MNIST} (Modified National Institute of Standards and Technology) dataset is a collection of handwritten digits commonly used to train image processing systems. For the MNIST classification result from Section \ref{sec:classification_tasks}, the original training dataset, $J$, comprises $N = 60000$ samples, wherein the fixed-proportion mixing parameters (for default numerical class ordering of digits from $1-10$) are:
\begin{equation*}
    \tilde{\alpha} = [0.0987, 0.1124, 0.0993, 0.1022, 0.0974, 0.0904, 0.0986, 0.1044, 0.0975, 0.0991]^T
\end{equation*}
The test dataset, $K$, comprises $N_{\text{test}} = 10000$ samples, with class proportions equivalent to the class proportions in the base MNIST test dataset. For MNIST reconstruction (see Section \ref{sec:reconstruction_tasks}), we utilize manual class imbalancing, reducing the number of samples comprising each numerical class $6-10$ by a factor of 5. The original training dataset, $J$, now contains $N = 36475$ samples, wherein the fixed-proportion mixing parameters (for default numerical class ordering of digits from $1-10$) are:
\begin{equation*}
    \tilde{\alpha} = [0.1624, 0.1848, 0.1633, 0.1681, 0.1602, 0.0297, 0.0324, 0.0344, 0.0321, 0.0326]^T
\end{equation*}
We note that the test dataset maintains the same class proportions as in the base MNIST test dataset. The features and labels within MNIST are summarized as follows:
\begin{itemize}
    \item Each feature (image) is of size $28 \times 28$, representing grayscale intensities from 0 to 255.
    \item Target Variable: The numerical class (digit) the image represents, ranging from 1 to 10.
\end{itemize}

\subsection{Fashion MNIST Dataset}
The \textbf{Fashion MNIST} dataset is a collection of clothing images commonly used to train image processing systems. For the Fashion MNIST classification result from Section \ref{sec:classification_tasks}, the original training dataset, $J$, consists of $N = 60000$ samples, wherein the fixed-proportion mixing parameters (for default numerical class ordering of clothing from $1-10$) are:
\begin{equation*}
    \tilde{\alpha} = [0.1, 0.1, 0.1, 0.1, 0.1, 0.1, 0.1, 0.1, 0.1, 0.1]^T = (0.1)\mathds{1}_{10}
\end{equation*}
The test dataset, $K$, comprises $N_{\text{test}} = 10000$ samples, with class proportions equivalent to the class proportions in the base Fashion MNIST test dataset. For Fashion MNIST reconstruction (see Section \ref{sec:reconstruction_tasks}), we use manual class imbalancing, reducing the number of samples within each numerical class $6-10$ by a factor of 5. The original training dataset $J$, now has $N = 36000$ samples. The fixed-proportion mixing parameters (for default numerical class ordering of clothing from $1-10$) are:
\begin{equation*}
    \tilde{\alpha} = [(0.1667)\mathds{1}_{5}^T, (0.0333)\mathds{1}_{5}^T]^T
\end{equation*}
We note that the test dataset maintains the same class proportions as in the base Fashion MNIST test dataset. The features and labels within Fashion MNIST are summarized as follows:
\begin{itemize}
    \item Each feature (image) is of size \(28 \times 28\), representing grayscale intensities from 0 to 255.
    \item Target Variable: The numerical class (clothing) the image represents, ranging from 1 to 10.
\end{itemize}

\subsection{CIFAR-10 Dataset}
The \textbf{CIFAR-10} dataset is a collection of color images categorized into $10$ different classes, and is commonly used to train image processing systems. For the CIFAR-10 classification result in Section \ref{sec:classification_tasks}, the original training dataset, $J$, comprises $N = 50000$ samples, wherein the fixed-proportion mixing parameters (for default numerical class ordering of categories from $1-10$) are:
\begin{equation*}
    \tilde{\alpha} = (0.1)\mathds{1}_{10}
\end{equation*}
The test dataset, $K$, comprises $N_{\text{test}} = 10000$ samples, with class proportions equivalent to the class proportions in the base CIFAR-10 test dataset. For CIFAR-10 reconstruction (see Section \ref{sec:reconstruction_tasks}), we use manual class imbalancing, reducing the number of samples in numerical classes $1-4, 7-10$ by a factor of 10. The original training dataset, $J$, now has $N = 14000$ samples. The fixed-proportion mixing parameters (for default numerical class ordering of categories from $1-10$) are:
\begin{equation*}
    \tilde{\alpha} = [(0.0357)\mathds{1}_{4}^T, (0.3571)\mathds{1}_{2}^T, (0.0357)\mathds{1}_{4}^T]^T
\end{equation*}
We note that the test dataset maintains the same class proportions found in the base CIFAR-10 test dataset. The features and labels within CIFAR-10 are summarized as follows:
\begin{itemize}
    \item Each feature (image) is of size $32 \times 32 \times 3$, with three color channels (RGB), and size 32 x 32 pixels for each channel, represented as a grayscale intensity from 0 to 255.
    \item Target Variable: The numerical class (category) the image represents, ranging from 1 to 10.
\end{itemize}

\subsection{Imagenette Dataset}
The \textbf{Imagenette} dataset contains a subset of $10$ classes from the ImageNet dataset of color images, and is commonly used to train image processing systems. The base Imagenette training dataset, $I$, comprises $N_I = 9469$ samples, and the base Imagenette test dataset, $K$, comprises $N_{\text{test}} = 3925$ samples. For the Imagenette classification result in Section \ref{sec:classification_tasks}, we utilize manual class imbalancing. Let $N_i \in \mathbb{N}$ be the number of samples in each class, $i \in \{1,\ldots,10 \}$, from $I$, where $\smash{N_I = \sum_{i=1}^{10} N_i}$. We define $\epsilon_i = 1 - 0.1i$, $\forall i \in \{1,\ldots,10\}$ as the linearly decreasing \textit{imbalance factor}. Accordingly, the original training dataset, $J$, has $\smash{N = \sum_{i=1}^{10} \epsilon_i N_i = 5207}$ samples. The fixed-proportion mixing parameters (for default numerical class ordering of categories from $1-10$) are:
\begin{equation*}
    \tilde{\alpha} = [0.1849, 0.1650, 0.1525, 0.1152, 0.1083, 0.0918, 0.0737, 0.0536, 0.0365, 0.0184]^T
\end{equation*}
We note that the test dataset maintains the same class proportions found in the base Imagenette test dataset. The features and labels within Imagenette are summarized as follows:
\begin{itemize}
    \item Each feature (image) is of size $224 \times 224 \times 3$, with three color channels (RGB), and size 224 x 224 pixels for each channel, represented as a grayscale intensity from 0 to 255.
    \item Target Variable: The numerical class (category) the image represents, ranging from 1 to 10.
\end{itemize}

\subsection{CIFAR-100 Dataset}
The \textbf{CIFAR-100} dataset is a collection of color images categorized into $100$ different classes, and is commonly used to train image processing systems. The base CIFAR-100 training dataset, $I$, has $N_I = 50000$ samples, and the base CIFAR-100 test dataset, $K$, has $N_{\text{test}} = 10000$ samples. For the CIFAR-100 classification result in Section \ref{sec:classification_tasks}, we utilize manual class imbalancing. Let $N_i \in \mathbb{N}$ be the number of samples in each class, $i \in \{1,\ldots,100 \}$, from $I$, whereby $\smash{N_I = \sum_{i=1}^{100} N_i}$. We define $\smash{\epsilon_i = 40^{-i/100}}$, $\forall i \in \{1,\ldots,100\}$ as the logarithmically decreasing \textit{imbalance factor}. Accordingly, the original training dataset, $J$, has $\smash{N = \sum_{i=1}^{100} \epsilon_i N_i = 13209}$ samples. The fixed-proportion mixing parameters (for default numerical class ordering of categories from $1-100$) are:
\begin{equation*}
    \tilde{\alpha} = [\tilde{\alpha}_1, \tilde{\alpha}_2, \ldots, \tilde{\alpha}_{100}]^T, \quad \text{where:} \quad \tilde{\alpha}_i = (\epsilon_i N_i) / N, \ \forall i \in \{1,\ldots,100 \}
\end{equation*}
We note that the test dataset maintains the same class proportions found in the base CIFAR-100 test dataset. The features and labels within CIFAR-100 are summarized as follows:
\begin{itemize}
    \item Each feature (image) is of size $32 \times 32 \times 3$, with three color channels (RGB), and size 32 x 32 pixels for each channel, represented as a grayscale intensity from 0 to 255.
    \item Target Variable: The numerical class (category) the image denotes, ranging from 1 to 100.
\end{itemize}

\subsection{IMDB Dataset}
The \textbf{IMDB} dataset is a collection of movie reviews, categorized as positive or negative in sentiment. We split the IMDB dataset such that the base IMDB training dataset, $I$, has $N_I = 40000$ samples, and the base IMDB test dataset, $K$, consists of $N_{\text{test}} = 10000$ samples. For the IMDB classification result in Section \ref{sec:classification_tasks}, we leverage manual class imbalancing, wherein numerical class $1$ retains $30\%$ of its samples. Accordingly, the original training dataset, $J$, has $N = 26000$ samples. The fixed-proportion mixing parameters (for default numerical class ordering of sentiment from $1,2$) are:
\begin{equation*}
    \tilde{\alpha} = [0.2307, 0.7693]^T
\end{equation*}
We note that the test dataset maintains the same class proportions as in the base IMDB test dataset. The features and labels within the IMDB dataset are summarized as follows:
\begin{itemize}
    \item Each feature (review) is tokenized and encoded as a sequence of word indices with a max length of 500 tokens. Sequences are padded or truncated to ensure uniform length.
    \item Target Variable: The numerical class (sentiment) the review represents, either 1 or 2.
\end{itemize}

\subsection{Mean Estimation Dataset}
The \textbf{Mean Estimation} dataset is a synthetic benchmark designed for regression tasks, wherein each example, $(x_j,y_j)$, comprises a 10-dimensional feature vector, $x_j$, of samples from one of four statistical distributions, and the mean, $y_j$, of this distribution. We create an imbalanced original training dataset, $J$, with $N=3000$ samples, where $J_1$ has $1000$ examples drawn from a normal distribution with $\sigma = 1$, $J_2$ has $1000$ examples drawn from an exponential distribution, $J_3$ has $800$ examples drawn from a chi-squared distribution, and $J_4$ has $200$ samples drawn from a uniform distribution. The fixed-proportion mixing parameters (for numerical ordering of distributions from $1-4$) are:
\begin{equation*}
    \tilde{\alpha} = [0.333, 0.333, 0.267, 0.067]^T
\end{equation*}
The test dataset, $K$, is created as a balanced dataset that has $1000$ examples from each distribution, wherein $N_{\text{test}} = 4000$. The Mean Estimation dataset features and labels are summarized as follows:

\begin{itemize} 
    \item Each feature (vector of samples) is generated from one of four statistical distributions (normal, exponential, chi-squared, uniform). The feature vectors are created by sampling from these distributions with means uniformly drawn from the interval $[0,1]$ for normal, exponential, and chi-squared distributions, and from $[20,50]$ for the uniform distribution.
     \item Target Variable: The mean parameter used to generate the vector of samples, representing the underlying expected value of the chosen distribution.
\end{itemize}

\subsection{Wine Quality Dataset}
The \textbf{Wine Quality} dataset consists of physicochemical tests on white and red wine samples, and the corresponding quality rating. We treat the wine type (white $=1$, red $=2$) as a categorical variable, wherein $k=2$. We split the Wine Quality dataset such that the base Wine Quality training dataset, $J$, has $N = 3248$ samples, and the base Wine Quality test dataset, $K$, has $N_{\text{test}} = 3249$ samples. For the Wine Quality regression result in Section \ref{sec:regression_tasks}, we utilize manual class imbalancing, reducing the number of samples in numerical class $1$ by a factor of 10. The original training dataset, $J$, now has $N = 1043$ samples, where the fixed-proportion mixing parameters (for numerical class ordering of wine type from $1,2$) are:
\begin{equation*}
    \tilde{\alpha} = [0.234, 0.766]^T
\end{equation*}
We note that the test dataset maintains the same class proportions as in the base Wine Quality test dataset. The features and labels within the Wine Quality dataset are summarized as follows:
\begin{itemize}
    \item Each feature (physicochemical tests) contains a set of test results, and is of size $11 \times 1$.
    \item Target Variable: The wine quality rating given to the set of physicochemical tests.
\end{itemize}

\subsection{California Housing Dataset}
The \textbf{California Housing} dataset contains housing data from California and their associated prices. As the ocean proximity variable is categorical ($<$1H OCEAN $=1$, INLAND $=2$, NEAR BAY $=3$, NEAR OCEAN $=4$), we denote $k=4$. We split the California Housing dataset such that the base California Housing training dataset, $J$, has $N = 10214$ samples, and the base California Housing test dataset, $K$, has $N_{\text{test}} = 10214$ samples. For the California Housing regression result in Section \ref{sec:regression_tasks}, we use manual class imbalancing, reducing the number of samples in numerical classes $1,2,4$ by a factor of 20. The original training dataset, $J$, now has $N = 3641$ samples. The fixed-proportion mixing parameters (for numerical class ordering of ocean proximity from $1-4$) are:
\begin{equation*}
    \tilde{\alpha} = [0.0615, 0.9055, 0.0154, 0.0176]^T
\end{equation*}
We note that the test dataset maintains the same class proportions as in the base California Housing test dataset. The features and labels in the California Housing dataset are summarized as follows:
\begin{itemize}
    \item Each feature (housing data) contains various housing attributes, and is of size $8 \times 1$.
    \item Target Variable: The housing price associated with the housing data.
\end{itemize}

\section{Experiment Details} \label{sec:experiment_details}

\subsection{Neural Network Architectures} \label{sec:network_architectures}
We provide comprehensive descriptions for six different neural network architectures designed for various tasks: classification, regression, and image reconstruction. Each of these architectures were employed to generate the respective empirical results pertaining to the aforementioned tasks.

\subsubsection{Fully Connected Networks}
We leverage fully connected networks in our analysis for regression on Mean Estimation, California Housing, and Wine Quality. The network consists of the following layers, wherein $d = 10$ for Mean Estimation, $d=11$ for Wine Quality, and $d = 8$ for California Housing:
\begin{itemize}
    \item \textbf{Fully Connected Layer (\texttt{fc1})}: Transforms the input features from a $d$-dimensional space to a $64$-dimensional space.
    \item \textbf{ReLU Activation (\texttt{relu})}: Applies the ReLU activation function to the output of \texttt{fc1}.
    \item \textbf{Fully Connected Layer (\texttt{fc2})}: Maps the $64$-dimensional representation from \texttt{relu} to a $1$-dimensional output.
\end{itemize}

\subsubsection{Convolutional Neural Networks}
We utilize the LeNet-5 convolutional neural network architecture in our analysis for image classification on MNIST and Fashion MNIST. The network consists of the following layers:
\begin{itemize}
    \item \textbf{Convolutional Layer (\texttt{conv1})}: Applies a 2D convolution with 1 input channel, 6 output channels, and a kernel size of 5.
    \item \textbf{ReLU Activation (\texttt{relu1})}: Applies the ReLU activation function to the output of \texttt{conv1}.
    \item \textbf{Max Pooling Layer (\texttt{pool1})}: Performs 2x2 max pooling on the output of \texttt{relu1}.
    \item \textbf{Convolutional Layer (\texttt{conv2})}: Applies a 2D convolution with 6 input channels, 16 output channels, and a kernel size of 5.
    \item \textbf{ReLU Activation (\texttt{relu2})}: Applies the ReLU activation function to the output of \texttt{conv2}.
    \item \textbf{Max Pooling Layer (\texttt{pool2})}: Performs 2x2 max pooling on the output of \texttt{relu2}.
    \item \textbf{Flatten Layer}: Reshapes the pooled feature maps into a 1D vector.
    \item \textbf{Fully Connected Layer (\texttt{fc1})}: Maps the flattened vector to a 120-dimensional space.
    \item \textbf{ReLU Activation (\texttt{relu3})}: Applies the ReLU activation function to the output of \texttt{fc1}.
    \item \textbf{Fully Connected Layer (\texttt{fc2})}: Maps the 120-dimensional input to a 84-dimensional space
    \item \textbf{ReLU Activation (\texttt{relu4})}: Applies the ReLU activation function to the output of \texttt{fc2}.
    \item \textbf{Fully Connected Layer (\texttt{fc3})}: Produces a 10-dimensional output for classification.
\end{itemize}

For image classification on CIFAR-10 and CIFAR-100, we employ an adapted, larger version of the LeNet-5 model, which we call `Large LeNet'. The network consists of the following layers, wherein $k = 10$ for CIFAR-10 and $k = 100$ for CIFAR-100.
\begin{itemize}
    \item \textbf{Convolutional Layer (\texttt{conv1})}: Applies 2D convolution with 3 input channels, 16 output channels, and a kernel size of 3.
    \item \textbf{ReLU Activation (\texttt{relu1})}: Applies the ReLU activation function to the output of \texttt{conv1}.
    \item \textbf{Max Pooling Layer (\texttt{pool1})}: Performs 2x2 max pooling on the output of \texttt{relu1}.
    \item \textbf{Convolutional Layer (\texttt{conv2})}: Applies 2D convolution with 16 input channels, 32 output channels, and a kernel size of 3.
    \item \textbf{ReLU Activation (\texttt{relu2})}: Applies the ReLU activation function to the output of \texttt{conv2}.
    \item \textbf{Max Pooling Layer (\texttt{pool2})}: Performs 2x2 max pooling on the output of \texttt{relu2}.
    \item \textbf{Convolutional Layer (\texttt{conv3})}: Applies 2D convolution with 32 input channels, 64 output channels, and a kernel size of 3.
    \item \textbf{ReLU Activation (\texttt{relu3})}: Applies the ReLU activation function to the output of \texttt{conv3}.
    \item \textbf{Max Pooling Layer (\texttt{pool3})}: Performs 2x2 max pooling on the output of \texttt{relu3}.
    \item \textbf{Flatten Layer}: Reshapes the pooled feature maps into a 1D vector of size $4 \times 4 \times 64$.
    \item \textbf{Fully Connected Layer (\texttt{fc1})}: Maps the flattened vector to a 500-dimensional space.
    \item \textbf{ReLU Activation (\texttt{relu4})}: Applies the ReLU activation function to the output of \texttt{fc1}.
    \item \textbf{Dropout Layer (\texttt{dropout1})}: Applies dropout with $p = 0.5$ to the output of \texttt{relu4}.
    \item \textbf{Fully Connected Layer (\texttt{fc2})}: Produces a $k$-dimensional output for classification.
\end{itemize}

\subsection{Mobile Neural Networks}
For image classification on CIFAR-10 and CIFAR-100, we also employ the MobileNet-V3 Small architecture. The network consists of the following layers, where $k = 10$ for CIFAR-10 and $k = 100$ for CIFAR-100.
\begin{itemize}
    \item \textbf{Convolutional Stem (\texttt{features0})}: 3 input channels, 16 output channels, kernel size 3, stride 2, padding 1, followed by BatchNorm and Hard-Swish activation.
    \item \textbf{Inverted Residual Block 1 (\texttt{features1})}: expansion factor 1, 16 to 16 channels, kernel size 3, stride 2, SE disabled, activation ReLU.
    \item \textbf{Inverted Residual Block 2 (\texttt{features2})}: expansion factor 4.5, 16 to 24 channels, kernel size 3, stride 2, SE disabled, activation ReLU.
    \item \textbf{Inverted Residual Block 3 (\texttt{features3})}: expansion factor 3.67, 24 to 24 channels, kernel size 3, stride 1, SE disabled, activation ReLU.
    \item \textbf{Inverted Residual Block 4 (\texttt{features4})}: expansion factor 4, 24 to 40 channels, kernel size 5, stride 2, SE enabled, activation Hard-Swish.
    \item \textbf{Inverted Residual Block 5 (\texttt{features5})}: expansion factor 6, 40 to 40 channels, kernel size 5, stride 1, SE enabled, activation Hard-Swish.
    \item \textbf{Inverted Residual Block 6 (\texttt{features6})}: expansion factor 6, 40 to 40 channels, kernel size 5, stride 1, SE enabled, activation Hard-Swish.
    \item \textbf{Inverted Residual Block 7 (\texttt{features7})}: expansion factor 3, 40 to 48 channels, kernel size 5, stride 1, SE enabled, activation Hard-Swish.
    \item \textbf{Inverted Residual Block 8 (\texttt{features8})}: expansion factor 3, 48 to 48 channels, kernel size 5, stride 1, SE enabled, activation Hard-Swish.
    \item \textbf{Inverted Residual Block 9 (\texttt{features9})}: expansion factor 6, 48 to 96 channels, kernel size 5, stride 2, SE enabled, activation Hard-Swish.
    \item \textbf{Inverted Residual Block 10 (\texttt{features10})}: expansion factor 6, 96 to 96 channels, kernel size 5, stride 1, SE enabled, activation Hard-Swish.
    \item \textbf{Inverted Residual Block 11 (\texttt{features11})}: expansion factor 6, 96 to 96 channels, kernel size 5, stride 1, SE enabled, activation Hard-Swish.
    \item \textbf{Convolutional Head (\texttt{features12})}: 1×1 Conv2d from 96 to 576 channels, followed by BatchNorm and Hard-Swish.
    \item \textbf{Adaptive Average Pooling (\texttt{features13})}: global average pool to 1×1.
    \item \textbf{Conv Head (\texttt{features14})}: 1×1 Conv2d from 576 to 1024 channels, followed by Hard-Swish.
    \item \textbf{Flatten Layer}: reshapes the 1024×1×1 tensor to a 1024-dimensional vector.
    \item \textbf{Fully Connected Layer (\texttt{classifier0})}: linear 1024 to 1024, followed by Hard-Swish.
    \item \textbf{Dropout Layer (\texttt{classifier2})}: dropout with $p=0.2$.
    \item \textbf{Fully Connected Layer (\texttt{classifier3})}: linear 1024 to $k$ for classification.
\end{itemize}

\subsubsection{Residual Neural Networks}
For image classification on Imagenette, we employ the ResNet-18 residual neural network architecture, which consists of the following layers:
\begin{itemize}
    \item \textbf{Convolutional Layer (\texttt{conv1})}: Applies a 7x7 convolution with 3 input channels, 64 output channels, and a stride of 2.
    \item \textbf{Batch Normalization (\texttt{bn1})}: Normalizes the output of \texttt{conv1}.
    \item \textbf{ReLU Activation (\texttt{relu})}: Applies the ReLU activation function to the output of \texttt{bn1}.
    \item \textbf{Max Pooling Layer (\texttt{maxpool})}: Performs 3x3 max pooling with a stride of 2 on the output of \texttt{relu}.
    \item \textbf{Residual Layer 1 (\texttt{layer1})}: Contains two residual blocks, each with 64 channels.
    \item \textbf{Residual Layer 2 (\texttt{layer2})}: Contains two residual blocks, each with 128 channels.
    \item \textbf{Residual Layer 3 (\texttt{layer3})}: Contains two residual blocks, each with 256 channels.
    \item \textbf{Residual Layer 4 (\texttt{layer4})}: Contains two residual blocks, each with 512 channels.
    \item \textbf{Average Pooling (\texttt{avgpool})}: Applies adaptive average pooling to reduce the spatial dimensions to 1x1.
    \item \textbf{Fully Connected Layer (\texttt{fc})}: Produces a 10-dimensional output for classification.
\end{itemize}

\subsubsection{Transformer Models}
For sentiment classification on IMDB Sentiment Analysis, we leverage a transformer architecture, which consists of the following layers:
\begin{itemize}
    \item \textbf{Embedding Layer (\texttt{embedding})}: Maps input tokens to 64-dimensional embeddings.
    \item \textbf{Positional Encoding (\texttt{pos\_encoder})}: Adds positional information to the embeddings with a maximum sequence length of 500.
    \item \textbf{Transformer Encoder (\texttt{transformer\_encoder})}: Applies a transformer encoder with 1 layer, 4 attention heads, and a hidden dimension of 128.
    \item \textbf{Pooling Layer (\texttt{pool})}: Averages the transformer outputs across the sequence length.
    \item \textbf{Dropout Layer (\texttt{dropout})}: Applies dropout with probability 0.1 to the pooled output.
    \item \textbf{Fully Connected Layer (\texttt{fc1})}: Maps the 64-dimensional pooled vector to 32-dimensional space.
    \item \textbf{ReLU Activation (\texttt{relu1})}: Applies the ReLU activation function to the output of \texttt{fc1}.
    \item \textbf{Fully Connected Layer (\texttt{fc2})}: Maps the 32-dimensional input to 2 output classes.
\end{itemize}

\subsubsection{Autoencoder Models}
For image reconstruction on MNIST, Fashion MNIST, and CIFAR-10, we employ an autoencoder. This network consists of the following layers, where $d = 784$ for MNIST and Fashion MNIST, and $d = 3072$ for CIFAR-10:
\begin{itemize}
    \item \textbf{Fully Connected Layer (\texttt{fc1})}: Transforms the input features from a $d$-dimensional space to a $128$-dimensional space.
    \item \textbf{ReLU Activation (\texttt{relu1})}: Applies the ReLU activation function to the output of \texttt{fc1}.
    \item \textbf{Fully Connected Layer (\texttt{fc2})}: Reduces the 128-dimensional representation to a 32-dimensional encoded vector.
    \item \textbf{Fully Connected Layer (\texttt{fc3})}: Expands the 32-dimensional encoded vector back to a 128-dimensional space.
    \item \textbf{ReLU Activation (\texttt{relu1})}: Applies the ReLU activation function to the output of \texttt{fc3}.
    \item \textbf{Fully Connected Layer (\texttt{fc4})}: Maps the 128-dimensional representation back to the original $d$-dimensional space.
    \item \textbf{Sigmoid Activation (\texttt{sigmoid1})}: Applies the Sigmoid activation function to ensure the output values are between 0 and 1.
\end{itemize}

\subsection{Focal Training} \label{sec:focal_training}
For the classification tasks outlined in Section \ref{sec:classification_tasks}, we compare learn2mix and classical training with focal loss-based neural network training (focal training). Let $\tilde{\alpha} \in [0,1]^k$ denote the vector of fixed-proportion mixing parameters, let $\mathcal{L}(\theta^t) \in \mathbb{R}^k$ denote the vector of class-wise cross entropy losses at time $t$, and let $\omega \in \mathbb{R}^k$ denote the vector of class-wise weighting factors, where $\forall i \in \{1,\ldots,k \}$:
\begin{align}
    \omega_i = \frac{[1 / (\tilde{\alpha}_i N)]}{\sum_{i'= 1}^k [1 / (\tilde{\alpha}_{i'} N)]} \times k.
\end{align}
The vector of predicted class-wise probabilities, $p \in [0,1]^k$, is given by $p = \exp{(-\mathcal{L}(\theta^t))}$, and we let $\Gamma \in \mathbb{R}_{\geq 0}$ be the focusing parameter. The focal loss at time $t$, $\mathcal{L}_{\text{FCL}}(\theta^t, \omega) \in \mathbb{R}_{\geq 0}$, is given by:
\begin{align}
    \mathcal{L}_{\text{FCL}}(\theta^t, \tilde{\alpha}) = \frac{1}{k} \sum_{i=1}^k (-\omega_i)(1 - p_i)^\Gamma \log(p_i).
\end{align}
Per the recommendations in \citep{lin2017focal}, we choose $\Gamma = 2$ in compiling the empirical results.

\subsection{SMOTE Training} \label{sec:smote_training}
For the classification tasks outlined in Section \ref{sec:classification_tasks}, we also compare learn2mix and classical training with neural networks trained on SMOTE-oversampled datasets (SMOTE training). Let $J$ denote the original training dataset, where the number of samples in each class, $i \in \{1,\ldots,k \}$ is given by $\tilde{\alpha}_i N$. After applying SMOTE oversampling, we obtain a new training dataset, $J^{\text{SMOTE}}$, with uniform class proportions, $\smash{\tilde{\alpha}^{\text{SMOTE}}_i = \frac{1}{k}}$, $\forall i \in \{1,\ldots,k\}$. The total number of samples in $J^{\text{SMOTE}}$, is given by:
\begin{align}
    N^{\text{SMOTE}} = \left( \max_{i \in \{1,\ldots,k\}} \tilde{\alpha}_i N \right) \times k.
\end{align}
In the original training dataset, $J$, we use a batch size of $M$, resulting in $P = N/M$ total batches. For consistency with learn2mix and classical training (see Section \ref{sec:classification_tasks}), we perform SMOTE training on $P$ batches of size $M$ from the SMOTE oversampled training dataset, $J^{\text{SMOTE}}$, during each epoch.

\subsection{IS Training} \label{sec:is_training}
For the classification tasks outlined in Section \ref{sec:classification_tasks}, we compare learn2mix and classical training with importance sampling–based neural network training (IS training) adapted from \citep{katharopoulos2018not} and \citep{johansson2022importance}. Let $J$ denote the original training dataset, and let $\mathcal{L}_{\mathrm{ind}}^{M}(\theta^t)\in\mathbb{R}^M$ denote the vector of individual cross‐entropy losses at time $t$ on a batch of size $M$ drawn uniformly from $J$.  We normalize these losses to sampling probabilities, $p_j \in [0,1]$, sample without replacement a subset of size $b = M/2$ according to $\{p_j\}$, and update the model by taking a gradient step on the average loss over that subset, where:
\begin{align}
  p_j = \frac{\mathcal{L}_{\mathrm{ind},j}^{M}(\theta^t)}{\sum_{j'=1}^M \mathcal{L}_{\mathrm{ind},j'}^{M}(\theta^t)}, \quad \text{and:} \quad 
\mathcal{L}_{\mathrm{IS}}(\theta^t) = \frac{1}{b}\sum_{r=1}^b \mathcal{L}_{\mathrm{ind},\,i_r}^{M}(\theta^t).
\end{align}

In the original training dataset $J$, we use a batch size of $M$, resulting in $P = N/M$ total batches, and perform IS training on $P$ batches of size $M$ during each epoch.

\subsection{CURR Training} \label{sec:curr_training}
For the classification tasks outlined in Section \ref{sec:classification_tasks}, we compare learn2mix and classical training with curriculum learning–based neural network training (CURR) following the self-taught scoring and fixed exponential pacing scheme of \citep{hacohen2019power}. Let $J$ be the original training dataset, and denote by $\tilde s_j = 1 - \hat p_j$ the self-taught score of sample $j$, where $\hat p_j$ is the network’s confidence in the correct label after preliminary convergence training on uniform mini-batches (this warm-up stage is used only to compute $\{\tilde s_j\}$ and is not included in our reported CURR timings, nor is any analogous stage required for learn2mix).  We sort the samples by increasing $\tilde s_j$ (easiest first) to obtain sorted indices $\{i_1,\dots,i_N\}$. At epoch $t$, let the curriculum fraction be:
\begin{align}
    \mathrm{frac}(t) = \min\bigl(\text{starting\_percent}\times \text{inc}^{\lfloor t/\text{step\_length}\rfloor},\,1.0\bigr),
\end{align}
with \(\text{starting\_percent}=0.5\), \(\text{inc}=1.2\), and \(\text{step\_length}=10\).  We form a curriculum subset of size \(\lfloor\mathrm{frac}(t)\,N\rfloor\) by taking the first \(\lfloor\mathrm{frac}(t)\,N\rfloor\) sorted indices, and train on mini-batches of size \(M\).  The curriculum loss at time $t$ is then:
\begin{align}
  \mathcal{L}_{\mathrm{CURR}}(\theta^t)
    &= \frac{1}{\lfloor\mathrm{frac}(t)\,N\rfloor}
      \sum_{r=1}^{\lfloor\mathrm{frac}(t)\,N\rfloor}
        \mathcal{L}_{\mathrm{ind},\,i_r}^{1}(\theta^t),
\end{align}
where \(\mathcal{L}_{\mathrm{ind},\,j}^{1}(\theta^t) \in \mathbb{R}\) is the individual cross-entropy loss on sample \(j\), and each epoch processes \(\lfloor\mathrm{frac}(t)\,N\rfloor / M\) batches of size \(M\).

\subsection{Neural Network Training Hyperparameters}
The relevant hyperparameters used to train the neural networks outlined in Section \ref{sec:network_architectures} are given in Table \ref{tab:hyperparameters}. All results presented in the main text were produced using these hyperparameter choices.

\begin{table*}[h]
\caption{Neural network training hyperparameters (grouped by task).}
\label{tab:hyperparameters}
\begin{center}
\begin{tabular}{c|c|c|c|c|c}
\hline 
\textbf{Dataset} & \textbf{Task} & \textbf{Optimizer} & \thead{\textbf{Learning} \\ \textbf{Rate} ($\boldsymbol{\eta}$)} & \thead{\textbf{Mixing Rate} ($\boldsymbol{\gamma}$) \\ (Learn2Mix)}  & \thead{\textbf{Batch} \\ \textbf{Size} ($\boldsymbol{M}$)} \\
\hline
MNIST & Classification & Adam & 0.0001 & 0.1 & 1000 \\
Fashion MNIST & Classification & Adam & 0.0001 & 0.5 & 1000 \\
CIFAR-10 & Classification & Adam & 7.0\text{e-}5 & 0.1 & 1000 \\
Imagenette & Classification & Adam & 1.0\text{e-}6 & 0.1 & 100 \\
CIFAR-100 & Classification & Adam & 0.0001 & 0.5 & 5000 \\
IMDB & Classification & Adam & 0.0001 & 0.1 & 500 \\
\hline
Mean Estimation & Regression & Adam & 5.0\text{e-}5 & 0.01 & 500 \\
Wine Quality & Regression & Adam & 0.0001 & 0.05 & 100 \\
California Housing & Regression & Adam & 5.0\text{e-}5 & 0.01 & 1000 \\
\hline
MNIST & Reconstruction & Adam & 0.0005 & 0.1 & 1000 \\
Fashion MNIST & Reconstruction & Adam & 1.0\text{e-}5 & 0.1 & 1000 \\
CIFAR-10 & Reconstruction & Adam & 1.0\text{e-}5 & 0.1 & 1000 \\
\hline
\end{tabular}
\end{center}
\end{table*}


\newpage
\section*{NeurIPS Paper Checklist}

\begin{enumerate}

\item {\bf Claims}
    \item[] Question: Do the main claims made in the abstract and introduction accurately reflect the paper's contributions and scope?
    \item[] Answer: \answerYes{} 
    \item[] Justification: The paper presents a new framework for accelerating neural network convergence. We provide comprehensive empirical results and theoretical guarantees to validate this claim.

\item {\bf Limitations}
    \item[] Question: Does the paper discuss the limitations of the work performed by the authors?
    \item[] Answer: \answerYes{} 
    \item[] Justification: The performance gains (large or limited) afforded by learn2mix are explicitly quantified in the empirical results section, and all methods, alongside ablation studies, are thoroughly discussed in the appendix.

\item {\bf Theory assumptions and proofs}
    \item[] Question: For each theoretical result, does the paper provide the full set of assumptions and a complete (and correct) proof?
    \item[] Answer: \answerYes{} 
    \item[] Justification: We provide theorems (with all relevant terms defined) in the main text, alongside comprehensive proofs in the appendix to verify the proposed theorems. 

    \item {\bf Experimental result reproducibility}
    \item[] Question: Does the paper fully disclose all the information needed to reproduce the main experimental results of the paper to the extent that it affects the main claims and/or conclusions of the paper (regardless of whether the code and data are provided or not)?
    \item[] Answer: \answerYes{} 
    \item[] Justification: Alongside the loss function and optimizer details presented in the main text, all neural network architectures and training hyperparameters are discussed and tabulated in the appendix.

\item {\bf Open access to data and code}
    \item[] Question: Does the paper provide open access to the data and code, with sufficient instructions to faithfully reproduce the main experimental results, as described in supplemental material?
    \item[] Answer: \answerYes{} 
    \item[] Justification: The complete code for reproducing all the empirical results for learn2mix are provided in the supplementary materials. All neural network architectures and training hyperparameters are also provided in the appendix.

\item {\bf Experimental setting/details}
    \item[] Question: Does the paper specify all the training and test details (e.g., data splits, hyperparameters, how they were chosen, type of optimizer, etc.) necessary to understand the results?
    \item[] Answer: \answerYes{} 
    \item[] Justification: The appendix details all neural network architectures utilized to generate the results presented in the main text and in the ablation studies. All hyperparameter choices are either explicitly specified in the main text/appendix, or tabulated in the appendix.

\item {\bf Experiment statistical significance}
    \item[] Question: Does the paper report error bars suitably and correctly defined or other appropriate information about the statistical significance of the experiments?
    \item[] Answer: \answerYes{} 
    \item[] Justification: All empirical results in the main text and the appendix include confidence intervals in the figures and tables to explicitly declare the statistical significance of all experiments.

\item {\bf Experiments compute resources}
    \item[] Question: For each experiment, does the paper provide sufficient information on the computer resources (type of compute workers, memory, time of execution) needed to reproduce the experiments?
    \item[] Answer: \answerYes{} 
    \item[] Justification: The compute resources (GPUs) used to produce the empirical results in the main text are specified in the empirical results section.
    
\item {\bf Code of ethics}
    \item[] Question: Does the research conducted in the paper conform, in every respect, with the NeurIPS Code of Ethics \url{https://neurips.cc/public/EthicsGuidelines}?
    \item[] Answer: \answerYes{} 
    \item[] Justification: The presented research conforms in all aspects with the NeurIPS Code of Ethics, and the authors have reviewed the NeurIPS Code of Ethics.

\item {\bf Broader impacts}
    \item[] Question: Does the paper discuss both potential positive societal impacts and negative societal impacts of the work performed?
    \item[] Answer: \answerYes{} 
    \item[] Justification: Accelerating neural network convergence in resource constrained regimes is an important capability to ensure fast and efficient neural network training --- the adoption of learn2mix can save compute cost and accelerate training. We find no negative societal impacts of our work. 
    
\item {\bf Safeguards}
    \item[] Question: Does the paper describe safeguards that have been put in place for responsible release of data or models that have a high risk for misuse (e.g., pretrained language models, image generators, or scraped datasets)?
    \item[] Answer: \answerYes{} 
    \item[] Justification: All the code is provided in the appendix and the final version will be maintained in a GitHub repository by the authors. The authors contact information will also be provided in the final version to prevent misuse.

\item {\bf Licenses for existing assets}
    \item[] Question: Are the creators or original owners of assets (e.g., code, data, models), used in the paper, properly credited and are the license and terms of use explicitly mentioned and properly respected?
    \item[] Answer: \answerYes{} 
    \item[] Justification: All relevant code and models used in the paper have been properly cited.

\item {\bf New assets}
    \item[] Question: Are new assets introduced in the paper well documented and is the documentation provided alongside the assets?
    \item[] Answer: \answerYes{} 
    \item[] Justification: The appendix contains comprehensive descriptions of all considered neural network architectures and modified datasets used to generate the empirical results.

\item {\bf Crowdsourcing and research with human subjects}
    \item[] Question: For crowdsourcing experiments and research with human subjects, does the paper include the full text of instructions given to participants and screenshots, if applicable, as well as details about compensation (if any)? 
    \item[] Answer: \answerNA{} 
    \item[] Justification: The paper does not involve crowdsourcing nor research with human subjects.

\item {\bf Institutional review board (IRB) approvals or equivalent for research with human subjects}
    \item[] Question: Does the paper describe potential risks incurred by study participants, whether such risks were disclosed to the subjects, and whether Institutional Review Board (IRB) approvals (or an equivalent approval/review based on the requirements of your country or institution) were obtained?
    \item[] Answer: \answerNA{} 
    \item[] Justification: The paper does not involve crowdsourcing nor research with human subjects.

\item {\bf Declaration of LLM usage}
    \item[] Question: Does the paper describe the usage of LLMs if it is an important, original, or non-standard component of the core methods in this research? Note that if the LLM is used only for writing, editing, or formatting purposes and does not impact the core methodology, scientific rigorousness, or originality of the research, declaration is not required.
    \item[] Answer: \answerNA{} 
    \item[] Justification: The core method development in this research does not involve LLMs as any important, original, or non-standard components.

\end{enumerate}

\end{document}